\documentclass{article}

\pdfoutput=1
\usepackage{arxiv}

\usepackage[utf8]{inputenc} % allow utf-8 input
\usepackage[T1]{fontenc}    % use 8-bit T1 fonts
\usepackage{hyperref}       % hyperlinks
\usepackage{url}            % simple URL typesetting
\usepackage{booktabs}       % professional-quality tables
\usepackage{amsfonts}       % blackboard math symbols
\usepackage{nicefrac}       % compact symbols for 1/2, etc.
\usepackage{microtype}      % microtypography
\usepackage{amsmath}
\usepackage{graphicx}
\usepackage{doi}
\usepackage{algorithm}
\usepackage{algpseudocode}
\usepackage{pgfplots}
\usepackage{filecontents}
\usepackage{subcaption}
\usepackage{tikz}
\usepackage{longtable}
\usepackage{siunitx}
\usepackage{ulem}
\pgfplotsset{compat=newest}

\newcommand\Bstrut{\rule[-1.2ex]{0pt}{0pt}}   % "bottom" strut
\newcommand{\pixelcount}[0]{pixel count}
\newcommand{\rsquared}[0]{$R^2$}

\newcommand{\nactions}[0]{$N_A$}

\newcommand{\pixelprobdistrib}[0]{$p(\textbf{y}_i|\textbf{x}_i)$}
\newcommand{\dinitial}[0]{$D_{\mathrm{L, init}}$}
\newcommand{\dlabelled}[0]{$D_{\mathrm{L}}$}
\newcommand{\dunlabelled}[0]{$D_{\mathrm{U}}$}

\newcommand{\cocotk}[0]{COCO10k}
\newcommand{\cocohk}[0]{COCO164k}
\newcommand{\ddqn}[0]{DDQN}
\newcommand{\dqn}[0]{DQN}
\newcommand{\classname}[1]{\textsc{#1}}

\newcommand{\themethod}[0]{MiSiCAL}

\title{Mining of Single-Class by Active Learning for Semantic Segmentation}

\author{
\href{https://orcid.org/0000-0002-2913-3937}{\includegraphics[scale=0.06]{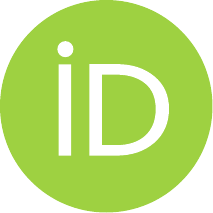}\hspace{1mm}Hugues
Lambert}\\
GSK.ai, GSK plc\\
N1C 4AG, London\\
<hugues.c.lambert@gsk.com>\\
	\And
	\href{https://orcid.org/0000-0002-2955-0669}{\includegraphics[scale=0.06]{orcid.pdf}\hspace{1mm}Emma
	Slade} \\
    GSK.ai, GSK plc\\
    N1C 4AG, London\\
    <emma.x.slade@gsk.com>\\
}

\renewcommand{\headeright}{}
\renewcommand{\undertitle}{}
\renewcommand{\shorttitle}{Mining of Single-Class by Active Learning for Semantic Segmentation}

\hypersetup{ pdftitle={Mining of Single-Class by Active Learning for Semantic Segmentation}, pdfsubject={q-bio.NC,
q-bio.QM}, pdfauthor={Hugues Lambert, Emma Slade}, pdfkeywords={Deep Q Learning,
Image Segmentation, Reinforcement Learning, Class Imbalance}, }

\begin{document}
\maketitle

\begin{abstract}
Several Active Learning (AL) policies require retraining a target model several
times in order to identify the most informative samples and rarely offer the
option to focus on the acquisition of samples from underrepresented classes.
Here the Mining of Single-Class by Active Learning (MiSiCAL) paradigm is
introduced where an AL policy is constructed through deep reinforcement learning
and exploits quantity-accuracy correlations to build datasets on
which high-performance models can be trained with regards to specific classes.
MiSiCAL is especially helpful in the case of very large batch sizes since
it does not require repeated model training sessions as is common in other AL
methods. This is thanks to its ability to exploit fixed representations of the
candidate data points. We find that MiSiCAL is able to outperform a random
policy on 150 out of 171 COCO10k classes, while the strongest baseline only
outperforms random on 101 classes.
\end{abstract}


\section{Introduction}
Semantic segmentation refers to the task of labelling images in a pixel-wise
fashion. Segmenting an image is often one of the first steps of image analysis
and provides information about the nature and localization of the content of an
image. High-performance image segmentation methods have enabled breakthroughs in
fields as varied as autonomous driving,\cite{treml2016speeding, kaymak2019brief}
where it can complement LIDAR and RADAR data to produce semantic depth
maps\cite{dickmann2016automotive} and biomedical image
segmentation\cite{ronneberger2015u} for surgery planning and
diagnosis.\cite{xiao2018weighted} In the task of brain tumour segmentation on
the BRATS dataset,\cite{menze2014multimodal} the application of deep learning
methods led to dramatic improvements from 2012\cite{ghaffari2019automated} as
convolutional neural networks (CNN) overtook random forest models in popularity
in 2015.

Recently, transformer-based architectures have achieved state-of-the-art
performance both on popular semantic segmentation
benchmarks\cite{chen2022vision,wang2022image} and medical
images\cite{valanarasu2021medical}. CNN-Transformer hybrids, in particular, have
also been used in the context of data scarce medical
settings.\cite{chen2021transunet, gao2021utnet} Original Vision Transformers
perform best when applied to large datasets and actually underperform CNNs on
smaller datasets, likely due to the inductive biases inherent to CNNs helping in
the scarce data regime at the cost of reduced flexibility and performance in
large data settings.\cite{dosovitskiy2020image} In contrast to early image
segmentation methods, for example based on colour clustering or
thresholding,\cite{zhang2006overview} which relied on very few parameters,
modern deep-learning models often possess millions of parameters requiring
training on large labelled training datasets to achieve their full potential.

While the amount of unlabelled data points available to a practitioner can be
very large, acquiring labels for these data points is typically a long, tedious
and expensive process. In the particular case of medical images, expert
annotators are often required, further complicating the labelling task. Larger
datasets usually result in lower model losses and greater
performance,\cite{henighan2020scaling} yet in practice, one is often limited by
a labelling budget, and it is often not feasible to label all available data
points. Determining a policy to select points that will lead to the highest
model performance for a fixed labelling budget is the crux of the active
learning (AL) paradigm. Pretraining models on self-supervised pretext
tasks\cite{chen2021empirical,caron2021emerging} and thus leveraging unlabelled data can
improve a model performance for a fixed labelling budget, yet labelled data are
often required to fine-tune the pretrained model and reach optimal
performance.\cite{tang2022self}

In theory, selecting points at random is likely to be a suboptimal policy as not
all points are expected to be equally informative.\cite{dasgupta2004analysis}
Paradoxically, the random policy is nonetheless a very strong
baseline in practice, especially for academic datasets.\cite{mittal2019parting,
hu2021towards} Several query policies have been developed leveraging model
uncertainty such as entropy sampling and Bayesian Active Learning by
Disagreement (BALD).\cite{gal2017deep, houlsby2011bayesian, holub2008entropy,
ash2019deep} in classification tasks, the \textit{coreset} method attempts to
select class diverse samples,\cite{sener2017active} while the batchBALD method
also promotes diversity within selected BALD batches by taking into account the
mutual information between the samples.\cite{kirsch2019batchbald} Another class
of policies aim to select unknown samples for labelling based on their expected
impact on the model considered. These methods include the Expected Model Change
Maximization (EMCM),\cite{cai2013maximizing} or the Influence Selection for
Active Learning (ISAL).\cite{liu2021influence}

Active Learning on Large Language Models (LLMs) such as BERT is very computationally
intensive.\cite{dor2020active} This means large batch sizes have to be used and
resources consumed to perform the training. In addition, there will be a wait
time for the annotator in between labelling steps while the model trains which
can be itself an issue.\cite{zhang2022survey} Full training at each iteration is
also impractical for pretrained large language
models.\cite{margatina2021importance} Therefore in practice, AL in the LLMs
context is limited to fine-tuning operations of a few epochs on small
models.\cite{kirk2022more}

Modern approaches include adversarial methods such as Variational Adversarial
Active Learning (VAAL),\cite{sinha2019variational} and Discriminative Active
Learning (DAL).\cite{gissin2019discriminative} Both VAAL and DAL train a binary
classifier on the samples representations in the latent space of a
variational autoencoder in the case of VAAL, or on the samples features as
extracted by a relevant model such as a CNN in the case of image classification
tasks for DAL. As such, neither VAAL nor DAL require repeated training of a
model other than their discriminator, which itself can be very lightweight, nor
explicit labelling (beyond marking a sample as ``labelled'') and therefore have
been shown to work well in large batch sizes. Another interesting work filters
sample candidates on their predicted Shapley values before selecting the most
promising based using the \textit{coreset} method.\cite{ghorbani2021data} Since the data
point Shapley value with regard to the task and model cannot be known before it
is labelled, the authors trained a Shapley value predictor on known data points.

While both VAAL and DAL partially circumvent retraining a model from scratch and
instead train lightweight classifiers on learnt latent representations of their
input data to guide the new sample acquisition process, updating the models able
to provide the latent representations of their original data can still be
computationally expensive and might even prove prohibitive in the case of
state-of-the-art large language models (LLMs). In addition, it is not obvious
how these methods could be adapted to build a dataset enriched in specific
classes using a binary classifier, since it might be hard to disentangle the
representations from labelled and unlabelled samples from a
single class from those of the other classes. Improving the performance metrics
with regard to a single class is often crucial as they might be bottlenecks
hard to optimise through general, class-agnostic AL frameworks.

Reinforcement learning based methods leverage features from other active
learning frameworks such as uncertainty and diversity sampling as inputs to a
Deep Q-Network (\dqn{}) aiming to directly optimize the model performance based on
the sample selection policy.\cite{slade2022deep, casanova2020reinforced} Using
reinforcement learning (RL) to directly learn an active learning policy--"learning
how to learn"--was introduced in the NLP literature and showed great
promise.\cite{fang2017learning} Such learned policies have been deployed in
active learning settings with humans in the loop for person
re-ID\cite{liu2019deep} and led to groundbreaking works in the case of ChatGPT,
a large language model based on InstructGPT.\cite{instructGPT} During the
training of InstructGPT, humans ranked up to 9 model outputs based on sample
prompts to train a reward model that is then used to fine-tune GPT-type models
to produce human-likeable outputs through a Proximal Policy Optimization (PPO)
algorithm.

Although the reward model is not used in a context of active learning
in the InstructGPT paper, it highlights the importance of obtaining timely
rewards during an iterative optimization process. In the original implementation
of BALD, for example, it is recommended to train a new model from scratch after
each batch selected during the active learning loop,\cite{gal2017deep} which is
often impractical. While this is generally wasteful, it becomes unfeasible for
large models and datasets. In addition, as models and datasets become large, the
marginal impact of a single data point becomes small, and it becomes difficult
to evaluate the individual contribution of newly added data
points.\cite{ghorbani2019data}

In reinforcement learning, rewards can be provided at the end of a training
episode to account for all the actions taken up to the end of the episode. Aside
from leading to reward sparsification, which is generally considered undesirable
in RL tasks,\cite{memarian2021self} it does not translate
well to the active learning field since there only is one ``episode'' in the sense
that each image can only be labelled once. Indeed, labelling the same data
points several times and ``forgetting'' them from one episode to the next would
lead to information leakage through the weights of the policy network and bias
the active learning procedure. A high convergence rate of the active learning
selection policy is crucial since every sample picked to improve the policy
depletes the labelling budget. Sample efficiency is therefore important to
successfully apply RL to active learning, and Deep Q
Networks are known to possess high sample efficiency thanks to their replay
buffer coming in multiple flavours.\cite{fedus2020revisiting,hessel2018rainbow}

Here, we leverage RL methods in the context of active learning to select image
patches from the
\cocotk{}\footnote[1]{\href{https://creativecommons.org/licenses/by/4.0/legalcode}{https://creativecommons.org/licenses/by/4.0/legalcode}}\cite{caesar2018coco}
dataset under the CC BY 4.0 licence. The image patches are selected by a \dqn{}
based on the features obtained from a pretrained segmentation model to optimise
the accuracy of that segmentation model on individual classes. The segmentation
model is not retrained nor fine-tuned in between selection events and only the
\dqn{} weights are updated based on the ground-truth semantic content of the
patches selected. This active learning paradigm is exceptionally lightweight as
the target model (here the semantic segmentation model) is not retrained at each
AL step, saving both training time but also the time necessary to update the
candidate data point features. The use of an RL framework also helps seamlessly
blend uncertainty and diversity in a single method, thereby bridging two popular
AL paradigms without the need to manually tune the importance of each aspect. We
envision this AL method might be extended to other classes of large deep
learning models, such as LLMs, which are impractical to retrain and therefore
ill-suited to traditional AL methods.

In contrast to most existing AL methods, the Mining of
Single-Class by Active Learning (\themethod{}), enables the selection of
datapoints from chosen classes instead of from all classes. This will be
particularly useful in the detection of outliers or out-of-distribution samples
which by definition belong to underrepresented classes and on which performance
metrics are difficult to improve.\cite{ali2019imbalance}
\section{Background}
\label{sec:background}
\subsection{Active learning}
\label{sec:active_learning_background}
The remit of AL is to provide policies to pick the most
informative samples from a pool of unlabelled data for which to query a label
from an oracle or a user. The objective is often to train models effectively
using as few data labels as possible, which is especially relevant when
acquiring labels is expensive, such as in medical settings. Active Learning
policies can be broadly divided between methods that use metrics computed
directly with the model meant to be used on the target task and those that
don't. DAL and VAAL belong to the latter category while uncertainty sampling,
ISAL and EMCM belong to the former. The \textit{coreset} method is perhaps the
most popular diversity-based sampling method.\cite{sener2017active} It attempts
to locate a set of samples $s^1$ with $i$ samples under a constrained annotation
budget $b$ that will minimize the maximum distance between the chosen samples
and the other $j$ samples in the dataset, that is:
\begin{align}
    \min_{s^1:|s^1| \leq b } \max_i \min_{j \in s^1 \cup s^0} \Delta(\boldsymbol{x_i}, \boldsymbol{x_j}),
\end{align}
with $s^0$ selected at random initially. The distance metric $\Delta(\cdot,
\cdot)$ is usually the Euclidean distance between the samples representations in
the model's latent space. The success of the \textit{coreset} method highlights
that using compressed representations of high-dimensional inputs such as images
as building blocks for AL heuristics can be very powerful. Seemingly simple
pixel class histograms have still been shown to be powerful
representations.\cite{casanova2020reinforced}
Uncertainty-based methods such as BALD and entropy are often used as benchmarks
due to their robustness, ease of implementation and, in the limit of small batch sizes,
very strong performances.
The Shannon entropy\cite{shannon1948mathematical} of the output of a
classification problem with $c$ classes given a training set $D_{\text{train}}$ can be
written as:\cite{gal2017deep}
\begin{align}
    \mathbb{H}[y=c|\textbf{x}, D_{\text{train}}] = -\sum_c p(y=c|\textbf{x}, D_{\text{train}}) \log p(y=c|\textbf{x}, D_{\text{train}}),
\end{align}
while BALD can be computed as:
\begin{align}
    \mathbb{I}[y,\boldsymbol{\omega}|\textbf{x}, D_{\text{train}}] = \mathbb{H}[y|\textbf{x}, D_{\text{train}}] - \mathbb{E}_{p(\boldsymbol{\omega}|D_{\text{train}})}[\mathbb{H}[y|\textbf{x}, \boldsymbol{\omega}]],
\end{align}
where $\boldsymbol{\omega}$ is the model parameters. With a distribution in model weights,
BALD attempts to select samples for which the average prediction is uncertain
but resulting from the average of highly certain predictions. These points are
expected to maximise the mutual information between the predictions and
the model posterior. Obtaining a weight distribution can be done by Monte Carlo
sampling T times dropout layers in a deep neural network, where BALD becomes
($p_c^t = p(y=c|\textbf{x}, D_{\text{train}})$ for the $t^{th}$ sampling of
$\boldsymbol{\omega}$):
\begin{align}
    \mathbb{I}[y,\boldsymbol{\omega}|\textbf{x}, D_{\text{train}}]= -\sum_c \Big(\frac{1}{T}\sum_t p_c^t\Big) \log \Big(\frac{1}{T}\sum_t p_c^t\Big) + \frac{1}{T}\sum_{c,t}p_c^t \log p_c^t.
    \label{eq:mc-bald}
\end{align}
As entropy and BALD information is computed at the pixel level for the output of
segmentation models, they may be pooled at the image or patch level to further
reduce their dimensionality. Both BALD information and a compressed
representation of an image sample, such as its latent representation or its
predicted pixel class histogram are powerful features by themselves. It is not
always obvious how much weight to give to each one to ensure the best samples
are picked. Passing this information as an observation to an agent following a
reinforcement learning algorithm can sidestep this issue and allow the agent to
find the proper weighting maximising the rewards it is receiving.

\subsection{Q-learning}
\label{sec:Q-learning_background}
An agent attempts to maximise the expected return it obtains over the long run,
with the return $G_t$ expressed as the sum of future rewards $R_t$, discounted
by a factor $\gamma$ as:\cite{sutton2018reinforcement}
\begin{align}
    G_t = R_t + \gamma R_{t+1} + \gamma^2 R_{t+2} + \gamma^3 R_{t+3} + ... = \sum_{k=0}^{\infty}\gamma^k R_{t+k}.
\end{align}
The return $G_t$ can be defined recursively by noting that the sum of future
expected rewards from step $t+1$ is $G_{t+1}$.
\begin{align}
    G_t & = R_t + \gamma R_{t+1} + \gamma^2 R_{t+2} + \gamma^3 R_{t+3} + ..., \\
        & = R_t + \gamma (R_{t+1} + \gamma R_{t+2} + \gamma^2 R_{t+3} + ...), \\
        & = R_t + \gamma G_{t+1}.
        \label{eq:rec_return}
\end{align}
In the context of a Markovian Decision process (MDP), the value function $v_\pi(s)$ of
a state $s$ following a policy $\pi$ is defined as the expected return when
starting from the state $s$ and following $\pi$ afterwards.
\begin{align}
    v_\pi(s) \doteq \mathbb{E}_\pi [G_t | S_t=s]  = \mathbb{E}_\pi\Bigg[ \sum_{k=0}^{\infty}\gamma^k R_{t+k}\Bigg| S_t=s \Bigg],
\end{align}
and similarly, the action-value function $q_\pi(s,a)$ associated with taking
action $a$ in state $s$ following $\pi$ is written:
\begin{align}
    q_\pi(s, a) \doteq \mathbb{E}_\pi [G_t | S_t=s, A_t = a ]  = \mathbb{E}_\pi\Bigg[ \sum_{k=0}^{\infty}\gamma^k R_{t+k} \Bigg| S_t=s, A_t = a \Bigg],
\end{align}
A policy having an expected return superior or equal to all other policies for
all states is referred to as an optimal policy. The optimal value function
$v_*(s)$ and action-value function $q_*(s,a)$ are described by the following
Bellman optimality equations, leveraging recursion from (\ref{eq:rec_return}):
\begin{align}
    v_*(s) &= \max_a \mathbb{E}_\pi \Big[R_{t} + \gamma v_*(S_{t+1}) \Big| S_t=s, A_t = a \Big], \label{eq:optimal_v}\\
    q_*(s, a) &= \mathbb{E}_\pi \Big[R_{t} + \gamma \max_{a'} q_*(S_{t+1}, a') \Big| S_t=s, A_t = a \Big].    \label{eq:optimal_q}
\end{align}
Equations (\ref{eq:optimal_v}) and (\ref{eq:optimal_q}) highlight that the value and
action-value functions at a given position in the MDP can be entirely described
in terms of the immediate reward received and the respective function value at
the next step. The functions $v_*(s)$ and $q_*(s, a)$ are unknown, and they are
usually estimated iteratively until convergence. The Temporal Difference method
provides a simple way to update current estimates V($s$) for a value function
$v_\pi(s)$ with a step size $\alpha$ as:
\begin{align}
    V(S_t) \leftarrow V(S_t) + \alpha \Big[ \underbrace{\underbrace{R_{t} + \gamma V(S_{t+1})}_{\mathrm{TD~target}} - V(S_t)}_{\mathrm{TD~error}}\Big].
    \label{eq:TD0}
\end{align}
This update is called TD(0) as the update uses only the current reward instead
of a taking into account future discounted rewards. The TD method is
bootstrapping as it is building an estimate upon a later estimate. The TD target
shown in (\ref{eq:TD0}) is the most recent estimate of V($S_t$) and both
converge towards the optimal state-value function over subsequent updates.

As an off-policy TD variation for control, the Q-learning builds estimates for the
action-value function $q_\pi$ independently of the policy followed and is
formulated as:
\begin{align}
    Q(S_t, A_t) \leftarrow Q(S_t, A_t) + \alpha \Big[ \underbrace{\underbrace{R_{t} + \gamma \max_a Q(S_{t+1}, a)}_{\mathrm{TD~target}} - Q(S_t, A_t)}_{\mathrm{TD~error}}\Big]
    \label{eq:q_network_update}
\end{align}
Approximating Q by a neural network forms the basic idea behind the original
\dqn{} implementation.
\section{Method}
\label{sec:Method}

\subsection{Overview}
\label{sub:overview}
Here we introduce the \themethod{} paradigm where a semantic segmentation model is
not trained between active sample selection events and its fixed sample
representations are leveraged by a RL-powered search procedure to acquire
samples of a chosen specific class. This is in contrast to popular AL policies
such as entropy or BALD which rely on successive training steps, at considerable
computational costs, to acquire promising samples. \themethod{} follows the
philosophy of modern methods such as DAL and VAAL which acquire samples in a
discriminative fashion based on fixed sample representations with a lightweight
classifier.

\themethod{} relies on the heuristic that increasing the representation of a
specific class in a training dataset is likely to improve the performance of a
model trained on this enriched dataset for metrics associated with the enriched
class. \themethod{} exploits the features of a pretrained segmentation model to
optimise the acquisition of image patches from the \cocotk{} set with the
objective of improving the performance of a semantic segmentation model on the IoU
metric of a single class. This is unusual as other AL policies generally attempt
to optimise the model performance over all classes simultaneously.

\begin{figure}[h!]
    \centering
    \includegraphics*[width=0.99\textwidth]{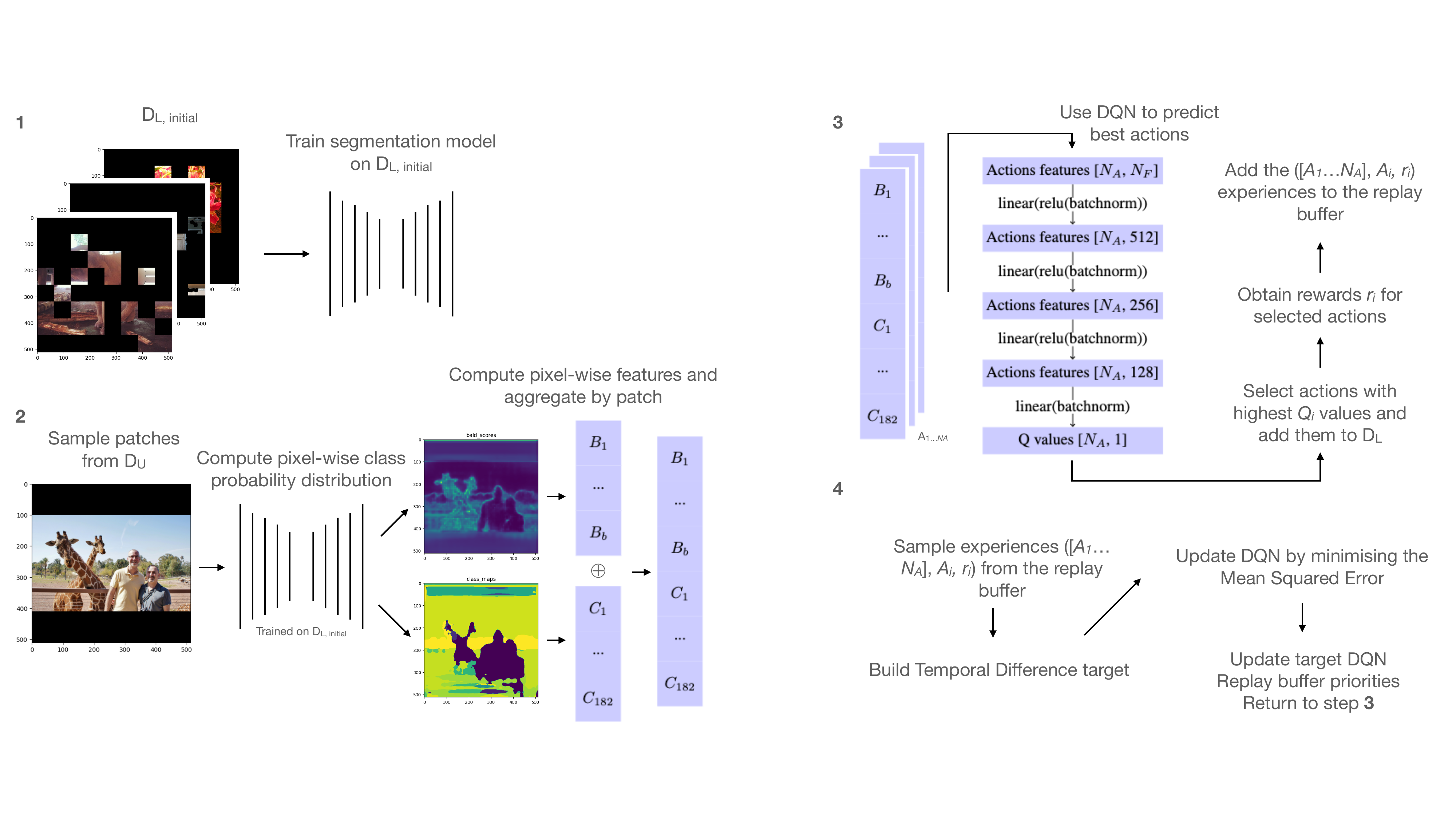}
    \caption{In \textbf{1}, an initial dataset \dinitial{} is used to train a
    segmentation model. \textbf{2} Calculation of the action features. An input
    image is resized, colour normalized, padded and its pixel wise class
    probability distribution \pixelprobdistrib{} is calculated via the
    segmentation model trained on \dinitial{}. The \pixelprobdistrib{}
    distribution is used to extract a pixel-wise BALD map. The map is then
    downsampled to the patch level using min, max and average pooling. Finally,
    a categorical vector is obtained from the segmented mask indicating whether
    a class is present in the patch. Both BALD and class features are then
    concatenated to produce the action features. As the segmentation model is
    not updated these action features are computed only once and kept fixed for
    the entire RL optimization procedure. In \textbf{3}, the \dqn{} is used to
    compute the Q values of a sample of \nactions{} from which the top-ranked
    are added to \dlabelled{}, the corresponding rewards are computed and the
    experience tuples are added to the replay buffer. Finally, in \textbf{4},
    the \dqn{} is updated using experiences sampled from the multistep
    prioritized experience replay through (\ref{eq:td_target}) and
    (\ref{eq:mse}). The target \dqn{} is then updated as well in
    (\ref{eq:soft_update}) along with the priorities of the replay buffer. A new
    action selection step is then performed in step \textbf{3}. Step \textbf{3}
    and \textbf{4} are repeated until a desired number of actions have been
    selected.
    }
    \label{fig:overview}
\end{figure}

The optimization procedure is outlined in Figure.~\ref{fig:overview}. A subset
of patches \dinitial{} corresponding to 2.5\% of \cocotk{} is randomly selected,
and a semantic segmentation model is trained on it. This trained segmentation
model is used to compute BALD features and class histograms for the remaining
624k patches of \cocotk{} (see section \ref{sub:implementation_details}). Patch
selection and \dqn{} updates are alternated until the annotation budget is
reached, in this case 5\% of the \cocotk{} dataset.

\begin{figure}[h!]
    \begin{center}
\begin{tikzpicture}
    \begin{axis}[
        ylabel={Cumulative \classname{skateboard} patches selected},
        xlabel={Selection event}
    ]
        \addplot[dashed, color=blue, mark size=2pt] table [x=Step, y=Value, col sep=comma]
        {explore_gamma_zero.csv};
        \addlegendentry{Exploring \dunlabelled{}}
        \addplot[dash dot, color=red, mark size=1pt] table [x=Step, y=Value, col sep=comma]
        {pretrain_gamma_zero.csv};
        \addlegendentry{Pretraining on \dinitial{}}
        % \draw[step=1cm,black,very thin, dashed] (-50,-50) grid (600, 500);
\end{axis}
\end{tikzpicture}
\caption{The \ddqn{} pretraining phase of an active learning loop aiming to
acquire patches containing \classname{skateboard} pixels (an example class) is shown
in red squares as the agent goes 4 times through \dinitial{} and becomes
increasingly proficient at picking patches containing \classname{skateboard} pixels.
After the pretraining, the agent quickly acquires more \classname{skateboard} pixels
from \dunlabelled{}.}
\label{fig:pretraing_explore}
\end{center}
\end{figure}
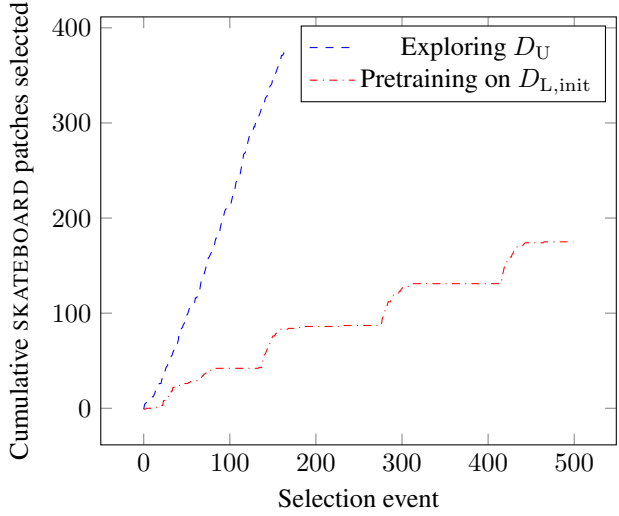

To start the patch selection process as effectively as possible, a Double Deep
Q-Network (\ddqn{})\cite{van2016deep} is first trained for 4 epochs on
\dinitial{} as shown in red squares on Figure.~\ref{fig:pretraing_explore} where
we show a representative plot for the \classname{skateboard}
class. At the end of each pretraining epoch and
going through \dinitial{} patches, the active learning procedure is restarted
from scratch while the weights and replay buffer of the \ddqn{}
are repeatedly updated.
During the later epochs of the pretraining, the agent quickly acquires all
\classname{skateboard} containing patches then is left to select patches without
\classname{skateboard} pixels. The agent is then left to explore \dunlabelled{}
with \dlabelled{} starting at \dinitial{} and the \ddqn{} weights copied from
the last pretraining step, as shown in blue triangles. In contrast to the
pretraining stage, the number of patches selected does not saturate as there are
many \classname{skateboard} images in \dunlabelled{} and the agent is not
running out of candidates.

One can notice that the patch acquisition curve follows an exponential before
plateauing rather than a step function even after training the \ddqn{}
in the pretraining phase. If the policy was fully greedy and
had access to all samples from the start to pick from, it is likely that a step
function would be observed. However, not only does the agent pick actions
randomly with a probability $\epsilon$ but when following the policy, it picks
only the top patches from a randomly selected subset which can contain only a
limited number of relevant patches.

\subsection{Implementation details}
\label{sub:implementation_details}
\paragraph*{Image preprocessing}
All images are resized to squares of 512$^2$ pixels with the largest dimension
resized to 512 pixels while the smallest dimension was padded with black to
reach 512 pixels. Patches measure 64$^2$ pixels unless stated otherwise. Labels
are resized in the same format as images with the padded area assigned a label
ignored in the loss function of the segmentation model. Pixel values were scaled to the
[0-1] range then normalized using a mean and standard deviation of respectively
(0.485, 0.456, 0.406) and (0.229, 0.224, 0.225). Training images come from the
\cocotk{} dataset while testing images are the 5k validation images from
\cocohk{}. \cocotk{} is therefore divided into 640k patches of 64$^2$ pixels as
10k images are resized to sizes of 512$^2$ then cut into 64 64$^2$ pixels patches.
\dinitial{} contains 2.5\% of \cocotk{} or 16k patches.

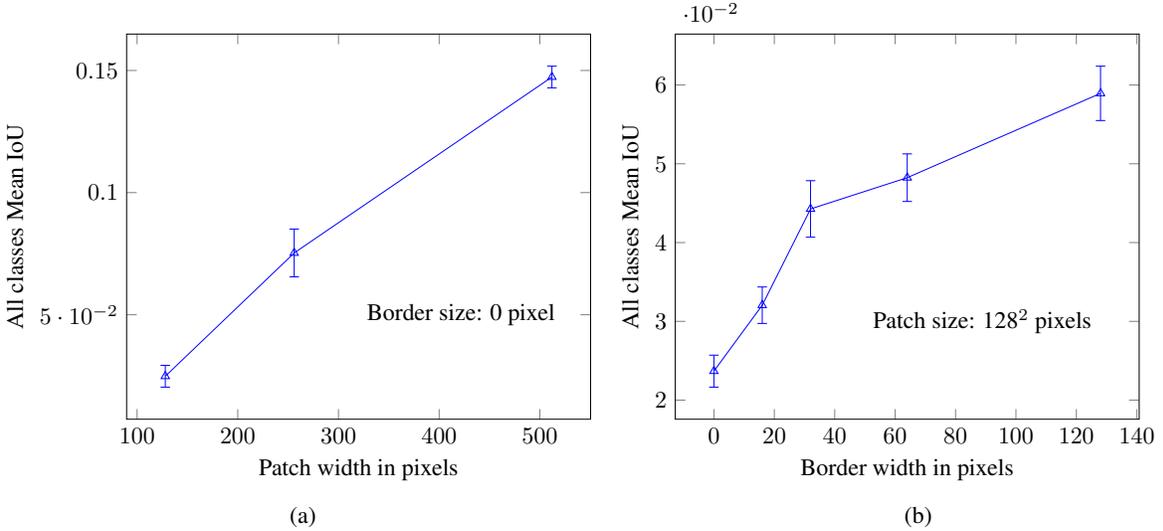
\begin{figure}[h!]
    \centering
    \begin{subfigure}[b]{0.49\textwidth}
        \begin{tikzpicture}[scale=0.9]
            \begin{axis}[
                xlabel={Patch width in pixels},
                ylabel={All classes Mean IoU}
                ]        \addplot [mark=triangle, color=blue, mark size=2pt]
                plot [error bars/.cd, y dir = both, y explicit]
                table [x=patchsize,y=MeanIoU, y error =IoUstddev] {patchsize.dat};
                \node [right] at (320,0.05) {Border size: 0 pixel};
        \end{axis}
        \end{tikzpicture}%
        \caption{}
        \label{fig:patch_size_accuracy}
    \end{subfigure}
\begin{subfigure}[b]{0.49\textwidth}
    \begin{tikzpicture}[scale=0.9]
        \begin{axis}[
            xlabel={Border width in pixels},
            ylabel={All classes Mean IoU}
            ]        \addplot [mark=triangle, color=blue, mark size=2pt]
            plot [error bars/.cd, y dir = both, y explicit]
            table [x=bordersize,y=MeanIoU, y error =IoUstddev] {bordersize.dat};
            \node [right] at (50,0.03) {Patch size: 128$^2$ pixels};
        \end{axis}
    \end{tikzpicture}%
    \caption{}
    \label{fig:border_size_accuracy}
\end{subfigure}
\caption[
    Evolution of the accuracy of the segmentation model with regard to the
    patch size
    ]{ Evolution of the accuracy of the segmentation model with regard to the
    patch size and patch border for a sample of 2,000 images from the \cocohk{}
    training dataset.}
\end{figure}

To test and compute the action features at inference time, no augmentation is
applied. During training, the following augmentation methods were used:
\begin{itemize}
    \item Random scaling using scaling factors of 0.5, 0.75, 1.0, 1.25, 1.5
    \item Random crop and padding to 512$^2$ pixels.
    \item Random horizontal flip with a probability of 0.5
    \item Either a RandomAutocontrast or a RandomEqualize transform as
    implemented by torchvision.\cite{torchvision}
\end{itemize}
Importantly, all inference is performed on full images and the subdivision into
patches is only performed later on the output pixel probabilities. Running the
inference on the individual patches directly prevents the segmentation model
from using the patch context and results in dramatic reduction in accuracy as
indicated in Figure.~\ref{fig:patch_size_accuracy}.

As shown in Figure.~\ref{fig:patch_size_accuracy}, it is interesting to see that
cutting large \cocohk{} images which are frequently 400-600 pixels wide in small
patches of 128$^2$ pixels results in a dramatic loss in accuracy
if patches are processed by themselves. Adding adjacent pixels,
without their labels, to small patches can alleviate the accuracy drop
(see Figure.~\ref{fig:border_size_accuracy}). Images are
augmented according to the procedure described in \ref{sub:augmentation}. Test
images are a selection of 2,000 \cocohk{} validation images and are resized to
the same size as the training images. The images are used whole and not cut into
patches. All runs are trained for a maximum of 50 epochs with a cosine annealing
learning rate schedule with a 50 epochs period and starting at 0.001 with a
minimum of 0. An early stopping criterion of the all class Mean IoU with
patience of 5 epochs. Training and test images are cut into patches with a
border made of pixels from the surrounding image to provide additional context.
The border pixels are ignored during the loss computation but indirectly
contribute to the training due to their presence in the convolution operations.
All patches from training images with their borders are resized to 321$^2$
pixels prior to training while testing images are also resized to 321$^2$ pixels
but not cut in patches. Error bars are the standard deviation of 5 seeds.
The image size itself has a large influence on the Mean IoU achievable on test
images of the same size as shown in Fig.~\ref{fig:img_size}. Indeed larger
images may be necessary to resolve small objects and contribute reach high
accuracies on the corresponding classes.
\paragraph*{Action representation}
The actions corresponding to individual patches are computed as shown in
Figure.~\ref{fig:overview}(\textbf{2}). The first part of the feature vector
consists of the maximum, minimum and mean of the BALD values \textbf{B} of all
$k$ pixels contained in the patch \textbf{x} considered as computed using (\ref{eq:mc-bald}).

\begin{align}
    \textbf{F}_B =\left\{
                \begin{array}{ll}
                  B_0 & : \max(\textbf{B}),\\
                  B_1 & : \min(\textbf{B}),\\
                  B_2 & : \mathrm{mean}(\textbf{B}).\\
                \end{array}
              \right.
\end{align}
The class feature vector is a categorical vector where the index $i$
corresponding to class $c_i$ is 1 if the segmentation model predicts at least
one pixel belonging to class $c_i$ in the patch and 0 otherwise. With
$p_k(y = c_i |\textbf{x})$ representing the probability of pixel $k$, part of patch
\textbf{x} to belong to class $c_i$, one can write the class feature vector
$\textbf{F}_{c_i}$ as:
\begin{align}
    \textbf{F}_{c_i} = 0\quad \mathrm{if}\quad \big[\max_c p_k(y = c |\textbf{x}) \neq p_k(y = c_i | \textbf{x}) \quad \forall \quad k\big] \quad \mathrm{else} \quad 1.
\end{align}

Eventually the action feature vector is obtained by concatenating the BALD and class feature vectors:
\begin{align}
    \textbf{F} = \textbf{F}_B \oplus \textbf{F}_{c_i} .
\end{align}

\paragraph*{Reward}
Once a patch has been selected, a reward is granted based on the semantic
content of the ground truth segmentation mask corresponding to that patch. The
reward is 1 if the patch contains at least a pixel from the class being
optimized for and 0 otherwise. Using the change in the class IoU
($\Delta$ IoU) after training for one epoch on the updated \dlabelled{} after
the addition of the selected samples didn't lead to a strong performance as
shown in Fig.~\ref{fig:rl_policies}. This is presumably because rewards cannot
be assigned at the patch level but need can only be obtained for the whole batch
of selected samples therefore providing more ``diluted'' information through
the rewards.

\paragraph*{Deep Q-networks}
The Q-values are approximated by Deep Q-learning , the \ddqn{}\cite{van2016deep}
flavour is used where the target network's weights $\theta_{\mathrm{target}}$
are updated at each local Q-network training batch with the local weights
$\theta_{\mathrm{local}}$ using a soft update of the
form:\cite{lillicrap2015continuous}
\begin{align}
    \theta_{\mathrm{target}} = (1-\beta) \theta_{\mathrm{target}} + \beta \theta_{\mathrm{local}},
    \label{eq:soft_update}
\end{align}
with $\beta$ set to 0.002. This helps avoid Q-values overestimations and
instabilities known to complicate the application of the \dqn{} method as shown
in Fig.~\ref{fig:target_update_freq}.

The local network is updated using a Bellman
equation,\cite{dietterich2000hierarchical} very similar to
(\ref{eq:q_network_update}) with the important difference that there is no
explicit state considered. Indeed, as there is no model update, the model state
does not change. Alternative state representations using the semantic content of
\dlabelled{} were attempted but did not lead to dramatic improvements. The
\ddqn{} update equation with learning rate $\alpha$ is adapted from
(\ref{sec:Q-learning_background}) with the difference that
actions are selected with two networks with identical structures but different
weights. The local network selects actions while the target network provides
their Q-values estimates:

\begin{align}
    Q^{\mathrm{local}}_{t+1}(S_t, A_t) = Q^{\mathrm{local}}_{t}(S_t, A_t)  + \alpha \Big[\underbrace{R_t + \gamma Q^{\mathrm{target}}_{t}\Big(S_t, \max_a Q^{\mathrm{local}}_{t}(S_{t+1}, a)\Big)}_{\mathrm{TD~target}} - Q^{\mathrm{local}}_{t}(S_t, A_t) \Big].
\end{align}

In the \ddqn{} case, the TD target is:
\begin{align}
    \mathrm{TD~target} = R_t + \gamma Q^{\mathrm{target}}_{t}\Big(S_t, \max_a Q^{\mathrm{local}}_{t}(S_{t+1}, A_{t})\Big),
    \label{eq:td_target}
\end{align}
while the \ddqn{} attempts to minimize the difference between the current Q-values
and their TD targets using the mean squared error loss:
\begin{align}
    \mathrm{MSE~Loss} = \Big(\mathrm{TD~target} - Q^{\mathrm{local}}_{t}(S_t, A_t) \Big)^{2} \quad .
    \label{eq:mse}
\end{align}
Indeed, the MSE Loss tends towards zero as the Q-values converge towards the optimal
$q_*(s,a)$ action-value function from (\ref{eq:optimal_q}).

\paragraph*{Deep Q-networks hyperparameters}
The rate of soft update $\beta$ is set to 0.002, setting $\beta$
to higher values leads to faster convergence of the $\theta_{\mathrm{target}}$
with $\theta_{\mathrm{local}}$ and ultimately leads to instabilities with
Q-values reaching very high ranges, which is an issue that the \ddqn{}
architecture is meant to avoid. On the other hand, setting $\beta$ too low slows
down the training as $\theta_{\mathrm{target}}$ become more insensitive to the
ongoing training of the local \dqn{}. \ddqn{} behaviours with both high and low
$\beta$ values are shown in Fig.~\ref{fig:target_update_freq}. Periodically
copying the weights $\theta_{\mathrm{local}}$ to $\theta_{\mathrm{target}}$
prevents the Q-values from diverging but leads to small scale instabilities
which prevents the \ddqn{} losses from converging towards very low values.
Gradients are clipped to 0.01 in an attempt to help and avoid instabilities
though it appears that the expected stabilisation is tiny at best as shown in
Fig.~\ref{fig:gradient_clipping}.

The network is trained with RMSProp with a learning rate of \num{1e-3}, a weight
decay of \num{1e-4} and a batch size of 256 experiences. The agent follows an
$\epsilon$-greedy policy with a constant $\epsilon$ value of \num{0.05}. Several
schedules of $\epsilon$-annealing were explored in Fig.~\ref{fig:eps_annealing}
with longer $\epsilon$-annealing generally performing better. Half of the
total labelling budget (2.5\% of \dunlabelled{}) is picked at random in
\dinitial{} and leveraged in pre-training, which corresponds to roughly the
same amount of randomly picked samples as following a linear $\epsilon$-annealing
over the selection of 5\% of \dunlabelled{}.

At each sampling event, a Q value is computed for a randomly
sampled subset of 2000 patches and the top 100 patches with the highest Q-values
are selected. Several discount factors $\gamma$ have been tried with the best
results obtained for $\gamma$=0, which was used to obtain the results presented.
Setting $\gamma$ to values closer to 1, encourages exploration and leads to
higher \dlabelled{} entropies as shown in Fig.~\ref{fig:gamma_long_eps} but it
seems that in the low data regime, focusing on short term rewards works best. In
Fig.~\ref{fig:rl_policies}, one can see that $\gamma=0$ outperforms larger
values in small batches, in particular when a longer $\epsilon$-annealing is
applied.

\paragraph*{Buffer}
To improve sample efficiency, many Q-learning methods use a replay buffer where
experiences and rewards are stored to update the \dqn{} at later stages instead
of discarding them after a \dqn{} update. When a replay buffer is present,
experiences are sampled at every step from the buffer instead of only consisting
of the last few experiences encountered by the agent. This helps the agent avoid
``forgetting'' about past experiences not recently encountered.
A multistep buffer builds rewards as a discounted sum of rewards from previous
steps and has been shown to lead to faster reward signal propagation and
ultimately faster learning.\cite{meng2021effect}

The multistep buffer records current rewards as an $n$-step
discounted accumulations of future rewards:
\begin{align}
    R^{(n)}_t = \sum_{k=0}^i \gamma^i R_{t+i}
\end{align}

The prioritized experience replay (PER)\cite{schaul2015prioritized} component
proved crucial to the effective training of the \ddqn{}. The prioritized
experience replay buffer assigns a higher weight to recent experiences and to
experiences for which the MSE Loss was large. This enables the agent to spend
more time learning on experiences for which its value function estimate was
leading to a high loss. This is useful for sparse rewards which otherwise would
contribute little to the \dqn{} training. Instead of sampling
the replay buffer uniformly, the PER samples an experience $i$ with probability
$P(i)$ written:
\begin{align}
    P(i) = \frac{p_i^\eta}{\sum_k p_k^\eta},
\end{align}
where $\eta$ is an exponent modulating the scale of the importance sampling,
reverting to uniform sampling for $\eta=0$. The individual $p_i$ can correspond
to the TD error from (\ref{eq:q_network_update}) plus a small number to
ensure a nonzero sampling probability for zero loss experiences. Alternatively
$p_i = \frac{1}{ \mathrm{rank(i)}}$ where $\mathrm{rank(i)}$ represents the rank
of experience $i$ when the replay buffer experiences are sorted by decreasing TD
errors. The former is used in our case. As the experience replay introduces bias
in the estimation of the expected values by changing the distribution of the
stochastic updates in an uncontrolled fashion, one can correct for this bias
using importance sampling weights $w_i$:
\begin{align}
    w_i = \Big(\frac{1}{N}\cdot \frac{1}{P(i)} \Big)^{\zeta}.
\end{align}
These weights multiply the TD error in the \dqn{} update (after scaling by the
maximum weight $w_j$). When $\zeta=1$, the non-uniform probabilities are fully
compensated for. In practice $\zeta$ is annealed from a starting value towards 1
which phases out the prioritization at the end of the training where unbiased
estimators are required. The replay buffer has a capacity of \num{1e5}
experiences. A multistep buffer\cite{hernandez2019understanding} with
prioritized experience replay is used with a step number of 3.

\paragraph*{Segmentation model}
\label{sub:augmentation}
The segmentation model architecture is LRASSP\cite{howard2019searching} as
implemented in Pytorch\cite{paszke2017automatic} and is optimized using
RMSprop\cite{hinton2012neural} with a \num{1e-3} learning rate and a cosine
annealing schedule with a period of 50 epochs, a weight decay of \num{1e-4} and
a momentum of \num{0.9}. An unweighted cross entropy loss is used. LRASSP is a
small model with 3.7M parameters which enables fast training and inference while
still reaching useful accuracies. In the low data regime, it was also found that
smaller models led to faster convergence. To enable the Monte-Carlo sampling of
the model outputs, two dropout layers with a probability of
50\%\cite{kirsch2019batchbald} were added to respectively the high- and low-
features before the output classifier layer.

\section{Results and Discussion}
\label{sec:Results_and_Discussion}
 A sample of results can be seen in Table.~\ref{tab:sample_classes_table} where
 \themethod{} is shown to perform well for classes starting with a high IoU
 metric such as \classname{bear} or \classname{horse}. In addition, patches
 containing classes for which the model used to calculate the features performs
 poorly, such as \classname{skis} and \classname{fork}, can be successfully
 acquired. This is likely due to strong correlations between classes that the
 \ddqn{} is able to exploit. For example, there are around 3k images containing
 \classname{skis} in the training dataset of \cocohk{}. In the majority of these
 images, \classname{skis} come in patches with \classname{person} and
 \classname{snow} classes. With the addition of the BALD features, it can be
 expected that the model is able to acquire \classname{skis} containing patches
 even if the class feature vector contains very few \classname{skis} labels. A
 full table can be found in the Appendix in Table.~\ref{tab:all_classes_table}.
The \themethod{} method outperforms random in 150 out of 171 classes, BALD
 outperforms in 101 out of 171 classes while the entropy policy outperforms in
 19 classes. \themethod{} is outperformed by random in a single class
 (\classname{clouds}). \themethod{} outperforms the other policies with
 statistical significance 65 times while BALD did so for a single class
 (\classname{scissors}) and neither random and entropy managed to outperform the
 other policies.

\begin{center}
    \begin{table}
        \caption{Comparison table of the end accuracies of models trained on 5\%
        of \cocotk{} for each class present in the dataset with the accuracies at
        \dinitial{} corresponding to 2.5\% of \cocotk{} shown for reference.
        Models are initially trained on a \dinitial{} corresponding to 2.5\% of
        the whole dataset then patches are acquired from \dunlabelled{} using
        features calculated from the models trained on \dinitial{} and added to
        \dlabelled{} until 5\% of \cocotk{} has been labelled. Intersection over
        Union figures are given in percentages plus or minus the standard
        deviation over 5 seeds. Outperformance against
        the other policies at 5\% are marked with an * and obtained using the
        Welch t-test for unequal variances with a p-value < 0.05 threshold. The
        largest mean accuracies are highlighted in bold.}
        \label{tab:sample_classes_table}
    \begin{tabular}{rrrrrr}
        \\
        Class &  Random 2.5\% & Random 5\% & Entropy 5\% & BALD 5\% &  \themethod{} 5\% \\ \hline \Bstrut \\
        Accuracy/baseball bat IoU &  0.02$\pm$0.04 & \mbox{0.00 $\pm$0.00\hspace{5pt}}&\mbox{0.15 $\pm$0.34\hspace{5pt}}&\mbox{0.02 $\pm$0.05\hspace{5pt}}&\textbf{5.30 $\pm$1.41}*\\
        Accuracy/bear IoU & 38.19$\pm$9.74 & \mbox{49.92$\pm$4.40\hspace{5pt}}&\mbox{37.53$\pm$2.79\hspace{5pt}}&\mbox{62.53$\pm$5.25\hspace{5pt}}&\mbox{\textbf{65.02$\pm$2.03\hspace{5pt}}}\\
        Accuracy/dog IoU & 31.39$\pm$3.86 & \mbox{35.85$\pm$2.10\hspace{5pt}}&\mbox{29.77$\pm$2.92\hspace{5pt}}&\mbox{40.65$\pm$2.12\hspace{5pt}}&\textbf{46.68$\pm$2.94}*\\
        Accuracy/fork IoU &  0.02$\pm$0.04 & \mbox{0.09 $\pm$0.19\hspace{5pt}}&\mbox{0.10 $\pm$0.21\hspace{5pt}}&\mbox{0.46 $\pm$0.47\hspace{5pt}}&\textbf{8.52 $\pm$1.34}*\\
        Accuracy/horse IoU & 30.17$\pm$3.93 & \mbox{36.97$\pm$3.73\hspace{5pt}}&\mbox{33.47$\pm$2.96\hspace{5pt}}&\mbox{35.01$\pm$2.94\hspace{5pt}}&\textbf{44.53$\pm$4.28}*\\
        Accuracy/hot dog IoU & 14.85$\pm$2.54 & \mbox{21.71$\pm$4.69\hspace{5pt}}&\mbox{17.86$\pm$4.90\hspace{5pt}}&\mbox{23.08$\pm$3.74\hspace{5pt}}&\textbf{28.67$\pm$2.12}*\\
        Accuracy/scissors IoU &  4.13$\pm$5.10 & \mbox{13.34$\pm$2.74\hspace{5pt}}&\mbox{7.08 $\pm$5.32\hspace{5pt}}&\textbf{24.25$\pm$5.72}*&\mbox{16.59$\pm$6.84\hspace{5pt}}\\
        Accuracy/skateboard IoU &  3.34$\pm$1.97 & \mbox{6.22 $\pm$1.88\hspace{5pt}}&\mbox{4.93 $\pm$2.53\hspace{5pt}}&\mbox{7.52 $\pm$3.71\hspace{5pt}}&\textbf{16.23$\pm$1.24}*\\
        Accuracy/skis IoU &  1.11$\pm$1.07 & \mbox{1.77 $\pm$1.42\hspace{5pt}}&\mbox{0.91 $\pm$0.92\hspace{5pt}}&\mbox{0.30 $\pm$0.56\hspace{5pt}}&\textbf{10.98$\pm$3.10}*\\

    \end{tabular}
\end{table}
\end{center}

From a certain perspective, it is remarkable that selectively accumulating
patches containing a target class and training on the acquired patches does
improve the class-specific performance metric. Indeed, in the correlations from
Figure.~\ref{fig:logpix_vs_accuracy_trends}, increasing a class pixel counts is
shown to improve a class-specific accuracy metric but in the context of pixels
from other classes also being present. One could imagine a case where pixel from
the \classname{bear} class could improve the accuracy of a model on the
\classname{dog} class at test time by reducing the model confusion. A model
having been trained exclusively on \classname{dog} images might mistake
\classname{bear} for \classname{dog} as they are both furry mammals. It is
assumed that the randomly acquired patches during the $\epsilon$-greedy
selection process help alleviate this potential problem.
In addition, it is known that features learnt from the majority classes
can be sufficient to classify under-represented classes.\cite{massi2022feature}

\paragraph*{Comparison of \themethod{} with other large batch size methods}
Going from a \dinitial{} of 16k patches to 32k patches effectively means
selecting the pixel equivalent of 250 full size images. batchBALD is a method
building on BALD aiming to maximise the intra-batch mutual information and has
been shown to be robust to large batch sizes.\cite{kirsch2019batchbald} However,
the batch size considered here is very large at 16k compared to the maximum
batch size of 40 considered in the paper, at which point batchBALD was showing
signs of a degrading performance and was essentially on par with a standard BALD
method with a batch size four times smaller.

\paragraph*{Patches}
An early design choice made was to process images as patches rather than whole.
This decision is not trivial as \cocotk{} image dimensions are in the range of a
few hundreds of pixels and could be processed directly whole or resized. It has
been reported however than labelling whole images at the pixel level was a
suboptimal strategy compared with spreading the same pixel budget over a larger
number of images.\cite{li2020uncertainty} Taking this reasoning to the extreme,
it was shown that a handful of pixels per image was sufficient to reach
accuracies comparable to those obtained with orders of magnitude more pixels
annotations spent on wholesale image annotations.\cite{shin2021all}

Interestingly, we found that annotating sparse regions within an image, such as
in patches, was crucial in reaching those accuracies as simply cutting down the
original size images into patches and processing them individually was
performing poorly. Indeed, one can see in Figure.~\ref{fig:patch_size_accuracy}
that cutting an image into increasingly smaller patches and training a model on
the individual patches leads to a quickly deteriorating model performance.
However, adding a border to the patches containing additional unlabelled image
pixels partially rescues the accuracy drop occurring due to using small patches
as shown in Figure~\ref{fig:border_size_accuracy}. Intuitively, this highlights
that context is critical in successfully classifying individual pixels, presumably
as information from pixels far away bring additional contextual information
through the successive convolution layers.

\begin{figure}
	\centering
	\includegraphics[width=0.45\textwidth]{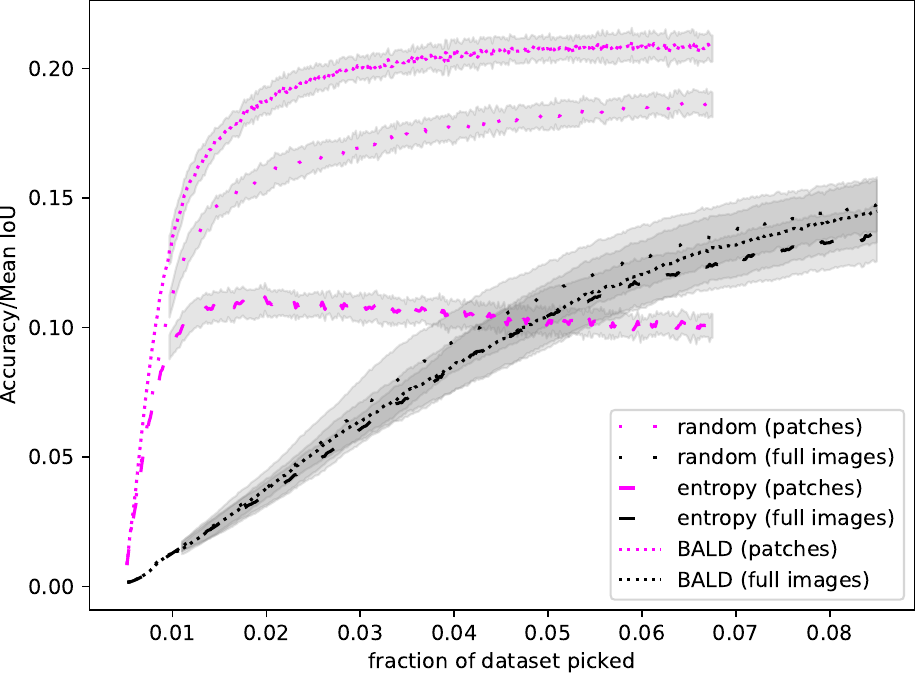}
	\caption{
        Comparison between actively selecting pixels to annotate by images or
        image patches for a fixed pixel annotation budget.
        }
	\label{fig:image_vs_patch}
\end{figure}

Beyond the improvements in model performance, we have also shown that performing
active learning from image regions can dramatically outperform the same
procedure when carried out on whole images. Figure.~\ref{fig:image_vs_patch}
reveals that BALD, Entropy and Random do not significantly differ when
considering whole images, while dramatic changes appear when selecting the
pixels by patches. Note that the entropy policy performs exceptionally poorly,
the reason being that images are resized and padded to a constant size before
being divided into patches. The black and unlabelled padded regions lead to high
model uncertainty which show up as high entropy zones even though there is no
semantic information in these patches. BALD does not suffer from this issue as
the high average uncertainty arises from an average of uncertain predictions
rather than an average of certain predictions. Although this could be alleviated
by discarding patches containing a lot of padding, this issue might not be known
a priori, so the poor performance of the entropy policy is left as is for
fairness. The acquisition process is started with 50 full images randomly
picked, and 2 additional images are selected from a randomly chosen subset of
2000 candidate images at each selection event. In the case of acquisition by
patches, 3200 patches were randomly selected, and 100 patches were labelled out
of a subset of 2000 patches randomly picked. Full images measure 512$^2$ pixels
while patches measure 64$^2$ pixels. The accuracies were measured on the 5k
images of the \cocohk{} test set. The LRASSP model is optimised using a batch
size of 16, while the learning rate is \num{1e-4} and the weight decay
\num{1e-4}. The model is trained for 1 epoch after each selection event and the
shaded area correspond to the standard deviation of 5 seeds. The BALD policy is
performed using 15 forward passes with the dropout layers enabled. The fraction
of dataset picked is measured in terms of pixel counts.

\paragraph{Heuristics for reward densification}
A crucial ingredient of any RL method is the reward
function. Since the Q-values are learnt from the rewards, the rewards are
essentially the only inputs of the model aside from the hyperparameters, and the
quality of the reward function can make or break a RL
method.

Previous works aiming to use RL in active learning in the context of semantic
segmentation directly used the segmentation model change in accuracy on a test
set as new labels were added to \dlabelled.\cite{casanova2020reinforced,
usmani2021reinforced} Directly optimizing on the model accuracy is a very
elegant approach as the end accuracy is the metric practitioners are actually
interested in, as opposed to, for example, minimizing the model uncertainty. One
issue is that the change in accuracy for the added labels does not scale well as
the marginal contributions of the acquired samples becomes vanishingly small as
\dlabelled{} becomes very large. The model accuracy can also be noisy and
dependent on the training parameters that were used. The change
in class accuracy $\Delta$ IoU was evaluated as a reward but did not lead to
dramatic improvements as shown in Fig.~\ref{fig:rl_policies}. Rewards can also
suffer from sparsity if the reward is provided after several selection steps or
at the end of a selection episode.

In practical active learning settings, there is no concept of an episode as is
customary in other RL contexts since data is only labelled once
(assuming exact ground truths can be acquired), and it is therefore very
important to quickly obtain a reliable reward signal, and avoid spending the
labelling budget learning on suboptimal candidates. Reward shaping and
densification aim to alleviate the issues associated with reward sparsity and
help speed up the learning process.\cite{gupta2022unpacking, memarian2021self}

Perhaps the simplest heuristic to improve a model accuracy in a classification
task is to increase the representation of the target class in its training
dataset. As shown in Figure.~\ref{fig:all_class_mean_iou_vs_subset_pix_count}
when randomly selecting 5 image subsets of size (118000, 60000, 30000, 15000,
7500, 3750, 1875, 937, 468), or a total of 45 subsets, from the 118k images of
the \cocohk{} training dataset and counting all pixels by class in each subset,
it appears that the mean class IoU scales almost linearly with the logarithm of
the pixel count. One can see that the pixel counts averaged by pixel bin in a
class-agnostic way displays a broadly linear relationship with the
class-agnostic average of IoUs in that bin. Trained on full images for 50 epochs
using LRASSP and RMSProp with a \num{1e-3} learning rate. The training was
stopped when the mean IoU stopped improving for 5 epochs. The cross entropy loss
is weighted with the inverse of the class frequency in the 118k \cocohk{}
training dataset multiplied by the minimum non-zero class frequency in the same
dataset.

Interestingly, the linear correlation broadly remains at the
class level as can be seen in Figure.~\ref{fig:logpix_vs_accuracy_trends}.
Indeed, both well represented classes, such as \classname{airplane} and underrepresented
classes, such as \classname{baseball bat} display a linear relationship between the
$\log_{10}$(\pixelcount{}) with the accuracy dropping very low below a 10k
pixels threshold. It would be tempting to use the logarithm of the pixel counts
as a proxy for the end IoU metric of the model simply based on the pixel
histogram of \dlabelled{}. With $i$ the class indices, $K{_i}$ a class specific
scaling factor and $h_i$ the pixel count of class $i$ in \dlabelled{}, one could
estimate the Mean IoU metric associated with a training dataset as:

\begin{align}
\mathrm{Mean~IoU} \approx \sum_i K_i \log(h_i).
\end{align}

In Figure.~\ref{fig:random_vs_uniform_sampling}, a thought experiment for
hypothetical datasets where the class IoUs indeed scale with the logarithm of
the pixel count and all $K_i$ factors are equal shows that for class imbalanced
datasets, selecting samples that would be expected to maximise the predicted
Mean IoU could strongly outperform the random sampling strategy. One can see
that when the pixel distribution sampled is balanced, the gap between random
sampling and sampling uniformly from all classes (which is expected to be a
strong baseline to build a high mean IoU) is small. On the other hand, for
larger class imbalances, the expected improvement from the random sampling is
larger. As the training set of \cocohk{} possesses a non-negligible degree of
class imbalance, picking samples based on the expected change in Mean IoU
induced by their addition to \dlabelled{} would be expected to strongly
outperform the random sampling strategy. Unfortunately the class specific
scaling coefficients $K_i$ are not known a priori and this policy barely
outperforms a random policy if equal $K_i$ are assumed.

As shown in Figure.~\ref{fig:histogram_of_pixel_iou_rsquared}, while the
relationship of $\log_{10}$(\pixelcount{}) with class IoUs are broadly linear,
the slope of these correlations have a wide distribution which does not
correlate with the class frequency, making it hard to estimate. Assuming equal
$K_i$ did outperform the random strategy in our tests (not shown) but did not
outperform BALD and therefore the focus was shifted from optimizing the model
performance on the IoU metric to optimizing the IoU metric of specific classes.
Interestingly, this distribution of $K_i$ means that directly optimising the
\dlabelled{} pixel count histogram entropy did not outperform BALD. Very high
pixel count histogram entropies can quickly be reached, but unfortunately this
does not translate to very high Mean IoU metrics. In datasets with very strong
class imbalances where a few classes dominate the pixel counts, such as
CityScapes,\cite{cordts2016cityscapes} directly optimising the entropy of the
\dlabelled{} normalised pixel histograms might be a viable option.

When a single class is considered, accumulating pixels from that class can
improve the IoU metric of that class. Both actions features and rewards are
categorical as using normalized pixel histograms was found to weaken the signal
from small objects such as knives and forks while using rewards linked to the
amount of pixels acquired tended to favour picking large surfaces which might
not have been informative or contained much context.

Since categorical annotations are used, where a vector simply
contains whether a class is present or not in the patch, it makes the
annotation process in real world settings more lightweight than if wholesale
pixel annotations were required. This requirement is softer than most AL methods
but more stringent than VAAL and DAL which only require to mark a sample as
labelled without the actual need to label it straight away. Compared to VAAL and DAL
which do not necessarily use features from a model specific task, \themethod{}
directly computes uncertainty metrics and representations from the task-specific
model. Associated with an RL-powered algorithm able to take advantage of user
defined rewards, the metrics can be leveraged online to reach complex objectives
beyond the ability to predict whether a sample has been labelled or not
irrespective of its semantic content.

\paragraph*{Shortcomings and limitations}
The \dqn{} powered procedure is not as easy to set up as the BALD or entropy
policies and may require some hyperparameters tuning. The
selection process is also dependent on the rewards function which rests on data
biases which might be specific to \cocohk{}. For datasets containing duplicates
or very similar images, there could be a risk of seeing the \dqn{} select such
images repeatedly to reap high rewards without actually leading to a high
performance training dataset. This issue would be present if searching through
highly redundant images such as those from a CCTV stream but would not be
expected to be problematic for highly diverse sources such as those from
histopathology slides or satellites images.

\section{Conclusion}
\label{sec:conclusion}
It is shown that Mining of Single-Class by Active Learning can serve as a
lightweight paradigm to dramatically boost the performance of a semantic
segmentation model on selected classes if convenient heuristics for data value
can be found. For \cocotk{}, the logarithm of the pixel count from a single
class in a training dataset was found to correlate with the IoU metric of
segmentation model trained on the dataset. This opens new possibilities to use
AL in the context of model optimization which are too costly or
impractical to repeatedly train to acquiring additional informative samples such
as LLMs. The possibility to acquire samples from a single class and subsequently
boost the accuracy of models trained with the new samples is also presented.
Within the \ddqn{} framework, a suitable weighting of model diversity and model
uncertainty emerges to allow the agent to maximise the acquisition of relevant
patches without the need to manually weight these contributions.
\bibliographystyle{unsrt}
\bibliography{main}

\clearpage
\appendix
\renewcommand{\thefigure}{A\arabic{figure}}
\renewcommand{\thetable}{A\arabic{table}}
\setcounter{figure}{0}
\setcounter{table}{0}

\section{Appendix}

\begin{figure}
	\centering
	\includegraphics[width=0.49\textwidth]{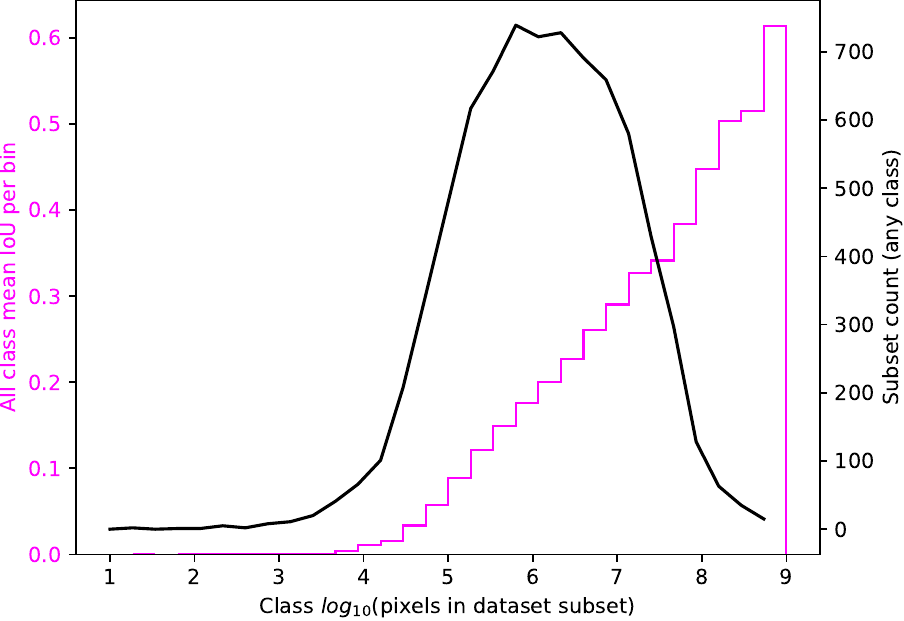}
	\caption{Class IoU binned and averaged by subset $\log_{10}$(\pixelcount)
	displays a stair-like behaviour in red with the y-axis on the left. The count
	of subset by class per bin is shown in green and displays a bell-shaped
	distribution with the y-axis displayed on the right. }
	\label{fig:all_class_mean_iou_vs_subset_pix_count}
\end{figure}

\begin{figure}
	\centering
	\includegraphics[width=0.49\textwidth]{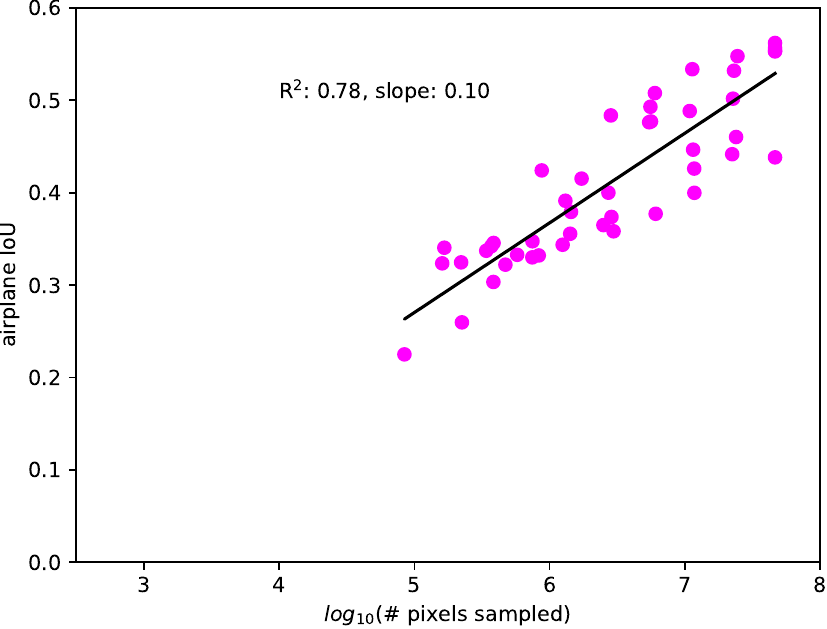}
	\includegraphics[width=0.49\textwidth]{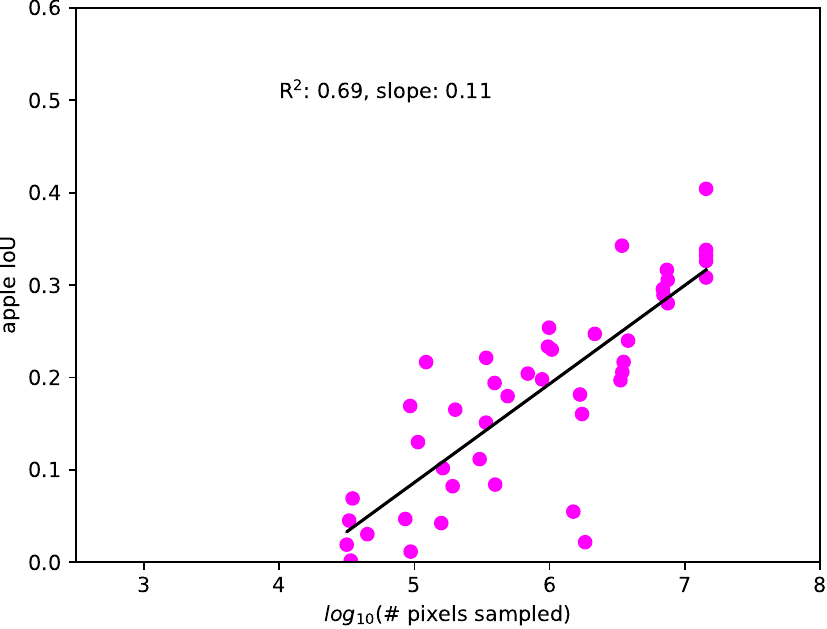}
	\includegraphics[width=0.49\textwidth]{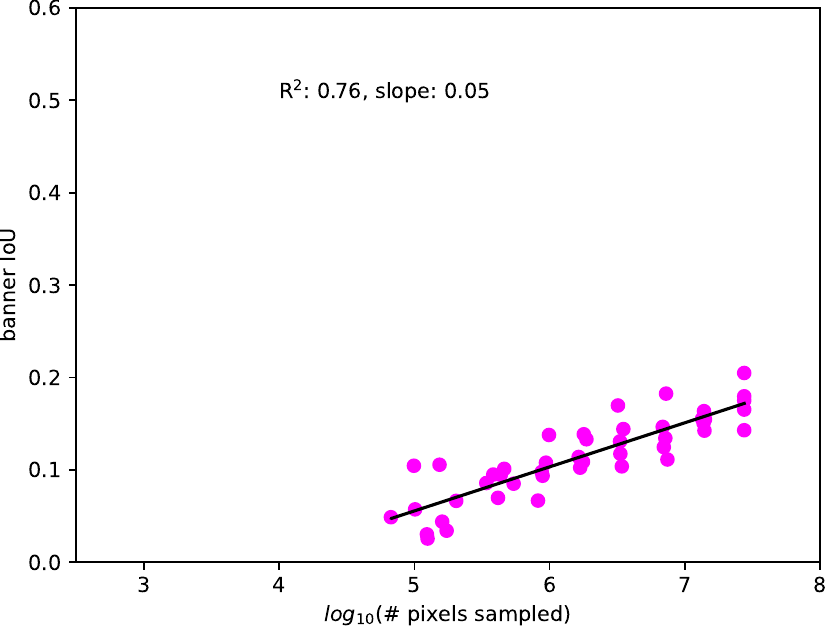}
	\includegraphics[width=0.49\textwidth]{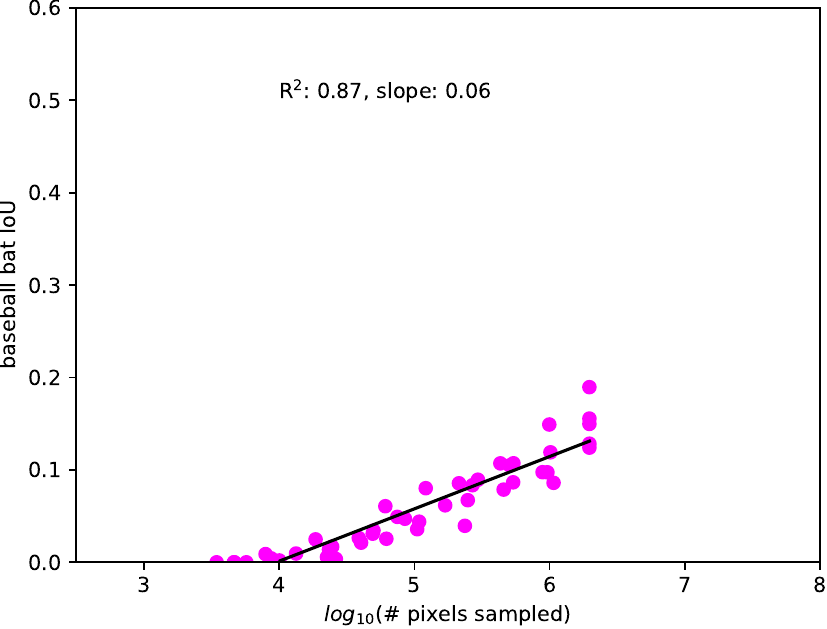}
	\caption{Relationship between the amount of pixels for a class in the
        training dataset versus the resulting class IoU. The relationship
        appears linear past a threshold of about $\log_{10}$(\pixelcount{}) = 4.5.
        }
	\label{fig:logpix_vs_accuracy_trends}
\end{figure}

\begin{figure}
	\centering
	\includegraphics[width=0.49\textwidth]{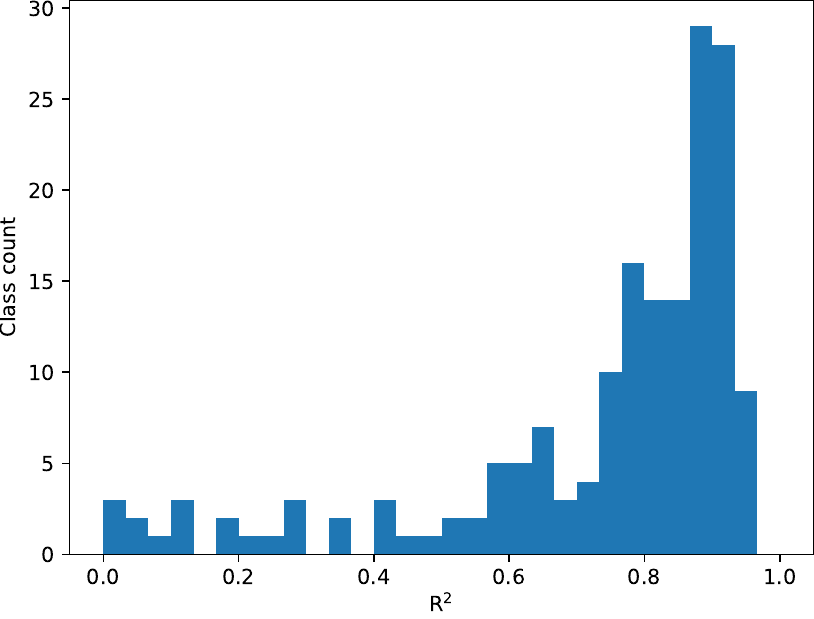}
	\includegraphics[width=0.49\textwidth]{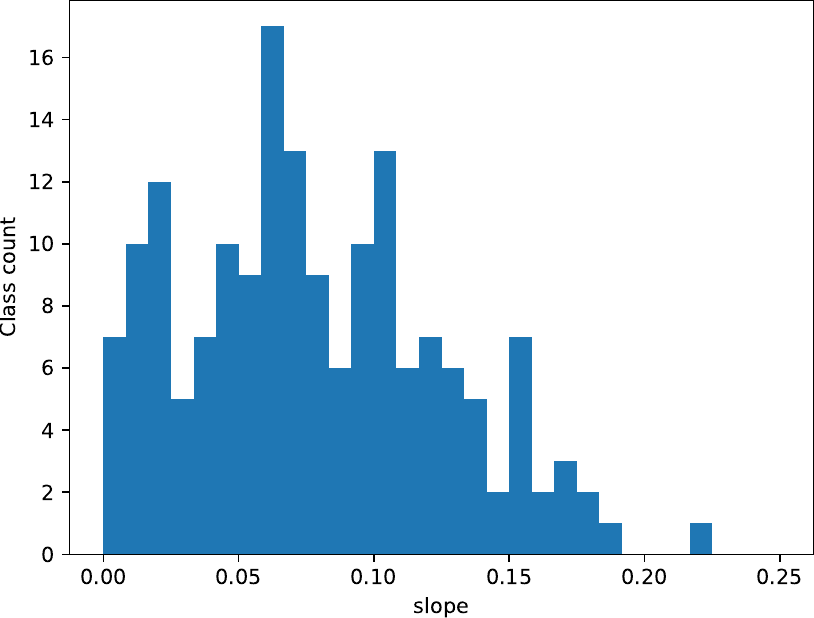}
	\includegraphics[width=0.49\textwidth]{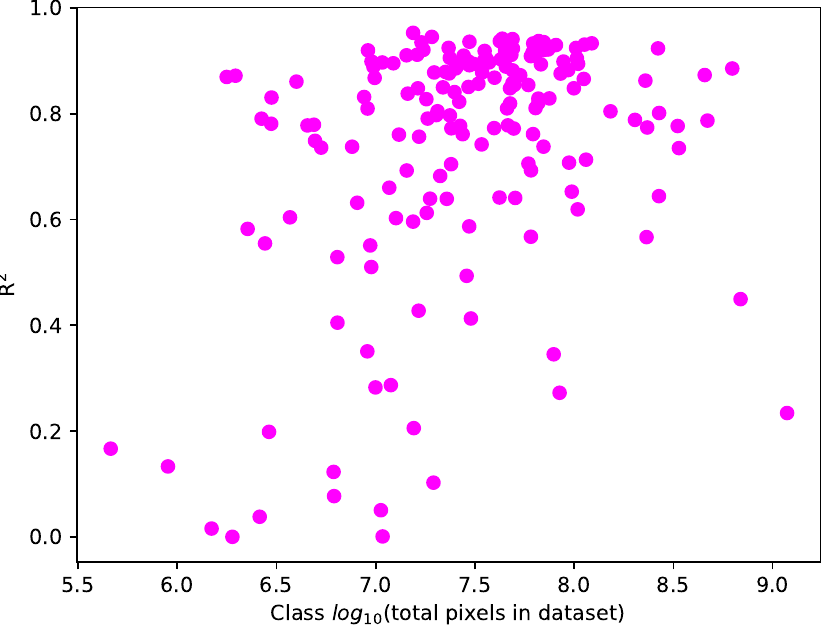}
	\includegraphics[width=0.49\textwidth]{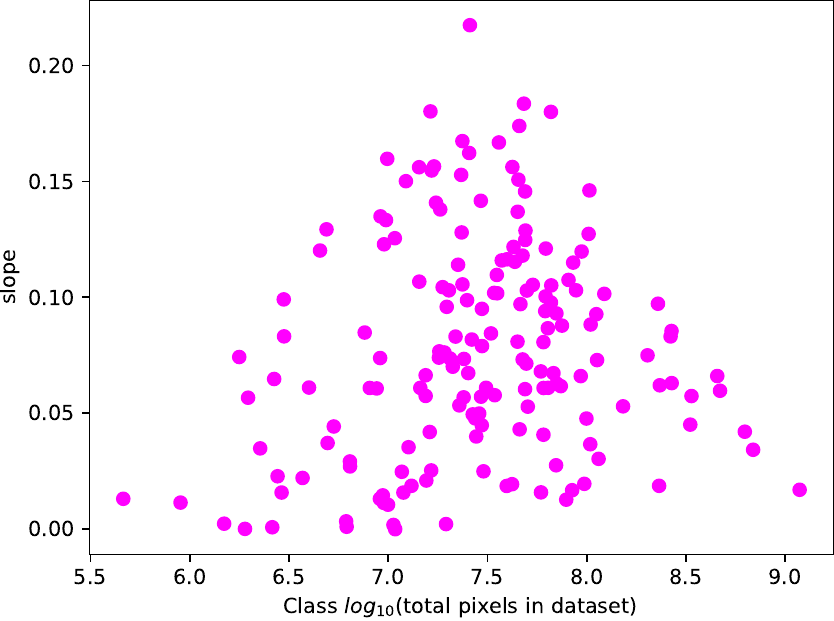}
	\caption{Histogram of the \rsquared{}, slopes values of the class pixel counts
        versus their IoU. It appears that over half of all classes exhibit a
        linear \pixelcount{} vs IoU correlation with a \rsquared{} value superior to
        0.8. On the other hand the value of the regression slopes are more
        uniformly spread in the 0 to 0.25 range. There does not seem to be a
        correlation between the total number of pixels from a class in the dataset
        with either \rsquared{} nor the slopes values.}
	\label{fig:histogram_of_pixel_iou_rsquared}
\end{figure}

\begin{figure}
    % This plot is in simulate samplling
	\centering
	\includegraphics[width=0.49\textwidth]{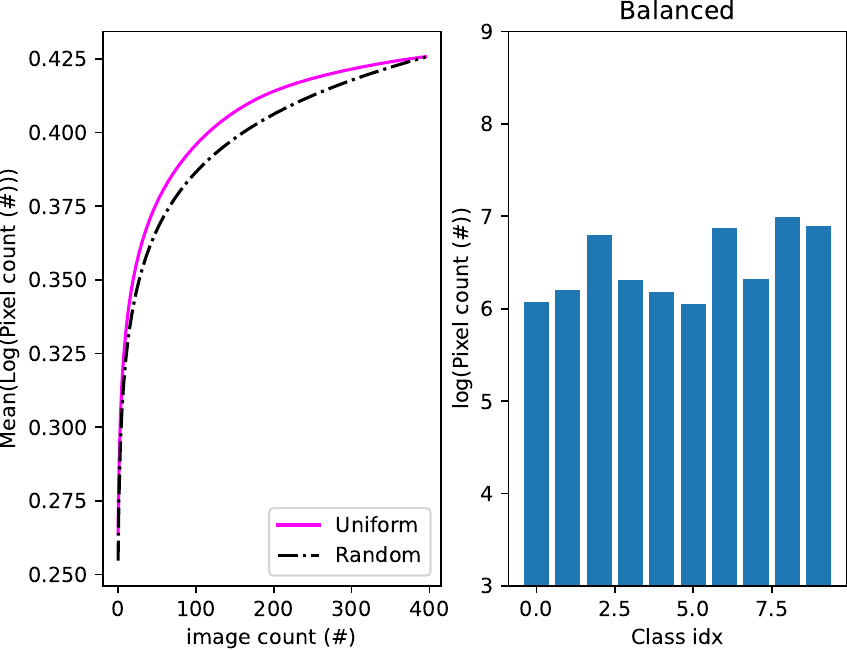}
	\includegraphics[width=0.49\textwidth]{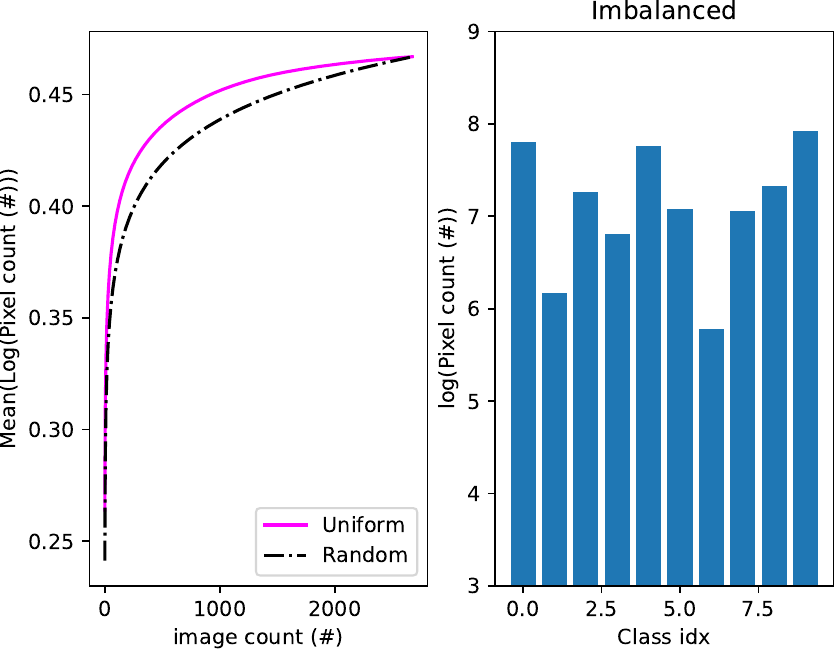}
    \parbox{\textwidth}{\vspace{1cm}}
    \includegraphics[width=0.49\textwidth]{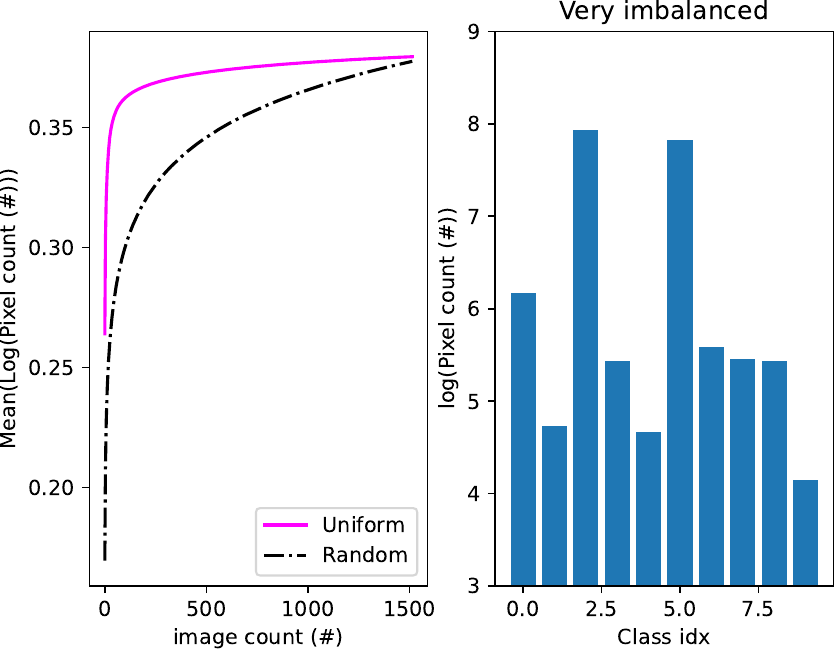}
	\includegraphics[width=0.49\textwidth]{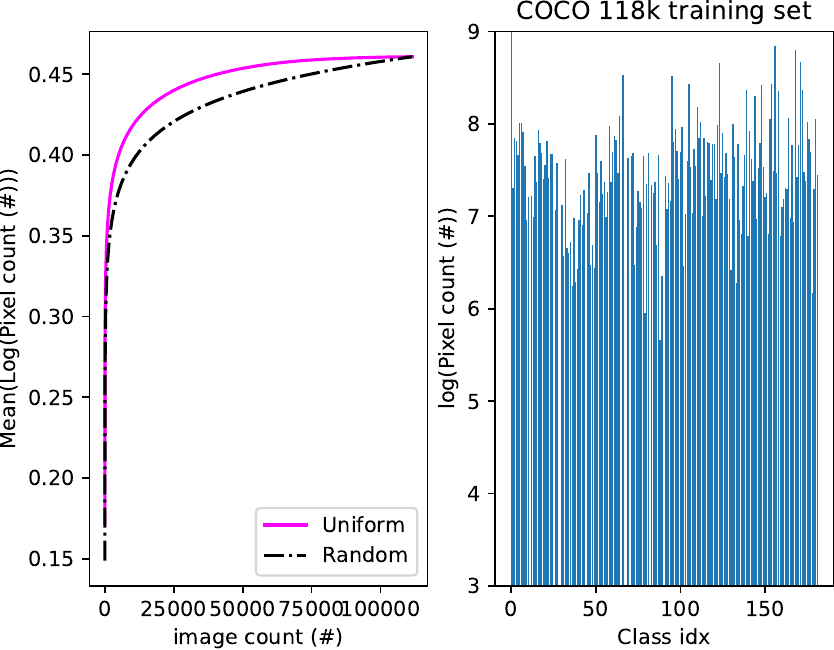}
	\caption{ Evolution of the mean IoU corresponding to a
        class uniform versus random sampling of a pixel
        distribution and assuming a linear relationship between the logarithm of
        the amount of pixels of a specific class in the training dataset and its
        IoU.}
	\label{fig:random_vs_uniform_sampling}
\end{figure}

\begin{figure}
    \centering
    \begin{tikzpicture}
        \begin{axis}[
            xlabel={Image width in pixels},
            ylabel={All classes Mean IoU}
            ]
            \addplot [mark=triangle, color=blue, mark size=2pt]
            plot [error bars/.cd, y dir = both, y explicit]
            table [x=cropsize,y=MeanIoU, y error=IoUstddev] {imagesize.dat};
        \end{axis}
    \end{tikzpicture}%
\caption[
    Evolution of the accuracy of the segmentation model with regard to the
    input image size ]{Evolution of the accuracy of the segmentation model with
    regard to the input image size for a sample of 10,000 images from the
    \cocohk{} training dataset. It appears that using larger images leads to
    higher end accuracies. Images are augmented according to the procedure
    described in \ref{sub:augmentation}. Test images are the 5,000 \cocohk{}
    validation images and are resized to the same size as the training images.
    The images are used whole and not cut into patches. All runs are trained for
    a maximum of 50 epochs with a cosine annealing learning rate schedule with a
    50 epochs period and starting at 0.001 with a minimum of 0. An early
    stopping criterion of the all class Mean IoU with patience of 5 epochs.
    Error bars are the standard deviation for 5 seeds.}
    \label{fig:img_size}
\end{figure}
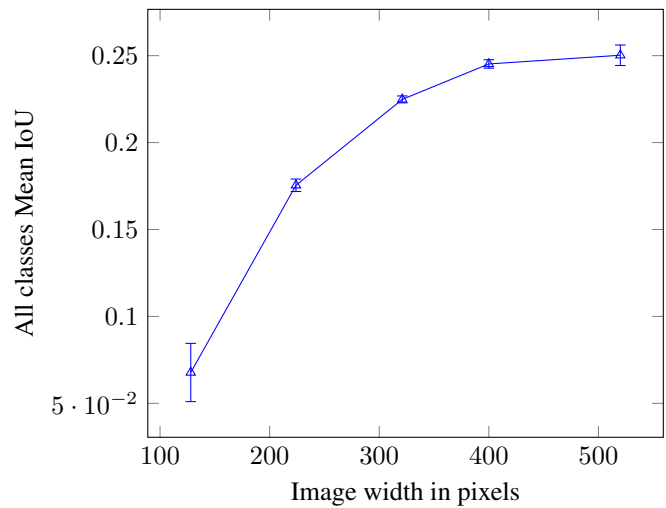

\begin{figure}
    \centering
    \begin{tikzpicture}
        \begin{axis}[
            xlabel={Selection events},
            ylabel={\classname{fork} cumulative reward},
            legend style={anchor=west}
            ]
            \addplot [thick, dotted, color=red, mark size=1pt]
            table [x=step,y=pawf_cr]  {gamma_500.dat};
            \addplot [thick, dashed, color=black, mark size=1pt]
            table [x=step,y=rgtk_cr]  {gamma_500.dat};
            \addplot [thick, color=green, mark size=1pt]
            table [x=step,y=dlgp_cr]  {gamma_500.dat};
            \addplot [thick, dash dot, color=blue, mark size=1pt]
            table [x=step,y=xcgt_cr]  {gamma_500.dat};
        \end{axis}
    \end{tikzpicture}%
    \begin{tikzpicture}
        \begin{axis}[
            xlabel={Selection events},
            ylabel={\dlabelled{} entropy},
            legend style={anchor=west}
            ]
            \addplot [thick, dotted, color=red, mark size=1pt]
            table [x=step,y=pawf_entr]  {gamma_500.dat};
            \addlegendentry{$\gamma=0.99$}
            \addplot [thick, dashed, color=black, mark size=1pt]
            table [x=step,y=rgtk_entr]  {gamma_500.dat};
            \addlegendentry{$\gamma=0.5$}
            \addplot [thick, color=green, mark size=1pt]
            table [x=step,y=dlgp_entr]  {gamma_500.dat};
            \addlegendentry{$\gamma=0.1$}
            \addplot [thick, dash dot, color=blue, mark size=1pt]
            table [x=step,y=xcgt_entr]  {gamma_500.dat};
            \addlegendentry{$\gamma=0$}
        \end{axis}
    \end{tikzpicture}%

    \begin{tikzpicture}
        \begin{axis}[
            xlabel={Selection events},
            ylabel={\classname{fork} IoU},
            legend style={anchor=west},
            ytick={0, 0.02, 0.04, 0.06, 0.08, 0.10},
            y tick label style={
                /pgf/number format/.cd,
                fixed,
                fixed zerofill,
                precision=2,
                /tikz/.cd
            }
            ]
            \addplot [thick, dotted, color=red, mark size=1pt]
            table [x=step,y=pawf_smoothed_acc]  {gamma_500.dat};
            \addplot [thick, dashed, color=black, mark size=1pt]
            table [x=step,y=rgtk_smoothed_acc]  {gamma_500.dat};
            \addplot [thick, color=green, mark size=1pt]
            table [x=step,y=dlgp_smoothed_acc]  {gamma_500.dat};
            \addplot [thick, dash dot, color=blue, mark size=1pt]
            table [x=step,y=xcgt_smoothed_acc]  {gamma_500.dat};
        \end{axis}
    \end{tikzpicture}%
    \begin{tikzpicture}
        \begin{axis}[
            xlabel={Selection events},
            ylabel={target \dqn{} loss},
            legend style={anchor=west},
            y tick label style={
                /pgf/number format/.cd,
                fixed,
                fixed zerofill,
                precision=1,
                /tikz/.cd
            }
            ]
            \addplot [thick, dotted, color=red, mark size=1pt]
            table [x=step,y=pawf_smooth_dqn_loss]  {gamma_500.dat};
            \addplot [thick, dashed, color=black, mark size=1pt]
            table [x=step,y=rgtk_smooth_dqn_loss]  {gamma_500.dat};
            \addplot [thick, color=green, mark size=1pt]
            table [x=step,y=dlgp_smooth_dqn_loss]  {gamma_500.dat};
            \addplot [thick, dash dot, color=blue, mark size=1pt]
            table [x=step,y=xcgt_smooth_dqn_loss]  {gamma_500.dat};
        \end{axis}
    \end{tikzpicture}%
\caption[]{
    Influence of the discount factor $\gamma$ on the behaviour of the \ddqn{}.
    At $\gamma=0$, the agent picks early on patches containing \classname{fork}
    pixels while at higher $\gamma$ values, more selection events are required to
    select the patches leading to rewards. This focus on the patches associated
    with the reward leads to a diminution of the entropy of the \dlabelled{}
    pixel histogram for smaller values of $\gamma$. For larger discount factors,
    the agent tends to pick patches containing more diverse pixels. The target
    \dqn{} loss is very stable for low discount factors and becomes
    progressively less so for higher values of $\gamma$. Interestingly, while
    picking the most patches and doing so earlier, the run with $\gamma=0$ does
    not lead to a much higher \classname{fork} IoU at the end of the selection
    procedure. This is presumably because the more diverse datasets picked by
    agents with a higher discount factor help the segmentation model to build
    better features. At each sampling event, 64 patches are selected from a
    random selection of 1280 patches, then added to \dlabelled{} and one epoch
    of training is performed on the whole \dlabelled{}. The process is repeated
    until 5\% of \dunlabelled{} (\cocotk{}) has been picked and only 250 patches
    are initially randomly sampled and added to \dlabelled{}. An $\epsilon$
    anneling is applied from 1 to 0.1 over 500 steps. The \ddqn{} batch size is
    set to 256 while the buffer contains 6400 experiences. Rewards are granted
    when the agent picks \classname{fork}-containing patches. The soft update is
    performed with $\beta=0.002$.
    }
    \label{fig:gamma_long_eps}
\end{figure}
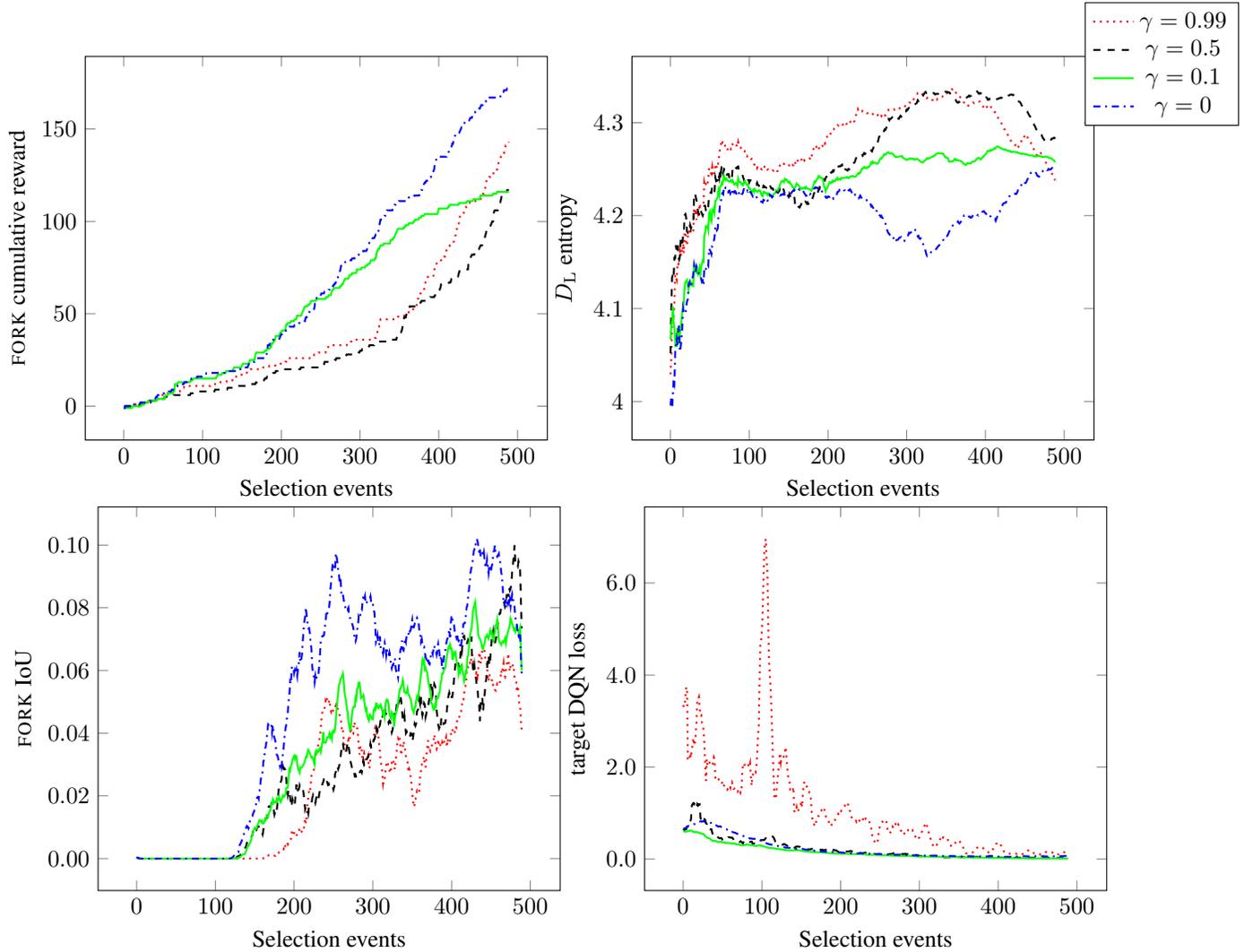

\begin{figure}
    \centering
    \begin{tikzpicture}
        \begin{axis}[
            xlabel={Selection events},
            ylabel={\classname{fork} cumulative reward},
            legend style={anchor=west}
            ]
            \addplot [thick, dotted, color=red, mark size=1pt]
            table [x=step,y=zugw_cr]  {target_freq_update.dat};
            \addplot [thick, dashed, color=black, mark size=1pt]
            table [x=step,y=rgyq_cr]  {target_freq_update.dat};
            \addplot [thick, color=green, mark size=1pt]
            table [x=step,y=yzet_cr]  {target_freq_update.dat};
        \end{axis}
    \end{tikzpicture}%
    \begin{tikzpicture}
        \begin{axis}[
            xlabel={Selection events},
            ylabel={\dlabelled{} entropy},
            legend style={anchor=west},
            ]
            \addplot [thick, dotted, color=red, mark size=1pt]
            table [x=step,y=zugw_entr]  {target_freq_update.dat};
            \addlegendentry{$\beta=0.002$}
            \addplot [thick, dashed, color=black, mark size=1pt]
            table [x=step,y=rgyq_entr]  {target_freq_update.dat};
            \addlegendentry{$\beta=0.02$}
            \addplot [thick, color=green, mark size=1pt]
            table [x=step,y=yzet_entr]  {target_freq_update.dat};
            \addlegendentry{$\beta=0.2$}
        \end{axis}
    \end{tikzpicture}%

    \begin{tikzpicture}
        \begin{axis}[
            xlabel={Selection events},
            ylabel={\classname{fork} IoU},
            legend style={anchor=west},
            ytick={0, 0.02, 0.04, 0.06, 0.08, 0.10},
            y tick label style={
                /pgf/number format/.cd,
                fixed,
                fixed zerofill,
                precision=2,
                /tikz/.cd
            }
            ]
            \addplot [thick, dotted, color=red, mark size=1pt]
            table [x=step,y=zugw_smoothed_acc]  {target_freq_update.dat};
            \addplot [thick, dashed, color=black, mark size=1pt]
            table [x=step,y=rgyq_smoothed_acc]  {target_freq_update.dat};
            \addplot [thick, color=green, mark size=1pt]
            table [x=step,y=yzet_smoothed_acc]  {target_freq_update.dat};
        \end{axis}
    \end{tikzpicture}%
    \begin{tikzpicture}
        \begin{axis}[
            xlabel={Selection events},
            ylabel={target \dqn{} loss},
            legend style={anchor=west},
            ymax=10
            ]
            \addplot [thick, dotted, color=red, mark size=1pt]
            table [x=step,y=zugw_smoothed_dqn_loss]  {target_freq_update.dat};
            \addplot [thick, dashed, color=black, mark size=1pt]
            table [x=step,y=rgyq_smoothed_dqn_loss]  {target_freq_update.dat};
            \addplot [thick, color=green, mark size=1pt]
            table [x=step,y=yzet_smoothed_dqn_loss]  {target_freq_update.dat};
        \end{axis}
    \end{tikzpicture}%
    \caption[adf]{
        A soft update carried out too aggressively with high $\beta$ values
        leads to instabilities, while a soft update too slow means the network
        is not updating its Q-value estimates fast enough and learns slower. The
        loss of the target \dqn{} in the case $\beta=0.2$ loss goes up to 300
        but this cropped on the image. At each sampling event, 64 patches are
        selected from a random selection of 1280 patches, then added to
        \dlabelled{} and one epoch of training is performed on the whole
        \dlabelled{}. The process is repeated until 5\% of \dunlabelled{}
        (\cocotk{}) has been picked and only 250 patches are initially randomly
        sampled and added to \dlabelled{}. An $\epsilon$ anneling is applied
        from 1 to 0.1 over 250 steps. The \ddqn{} batch size is set to 256 while
        the buffer contains 6400 experiences. The discount factor $\gamma$ is
        set to 0.99. Rewards are granted when the agent picks
        \classname{fork}-containing patches. Gradients are clipped to 0.01.
        }
        \label{fig:target_update_freq}
    \end{figure}
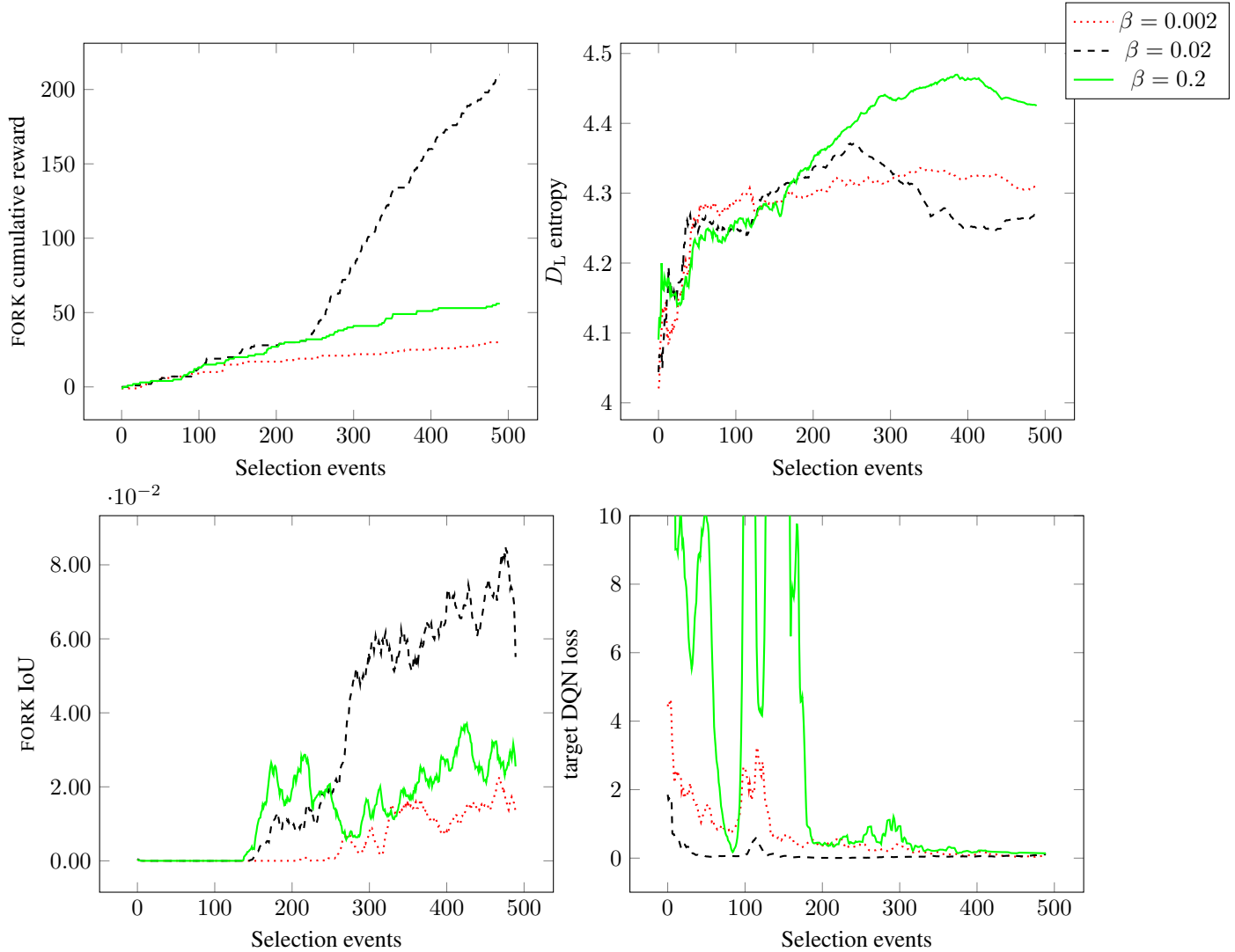

    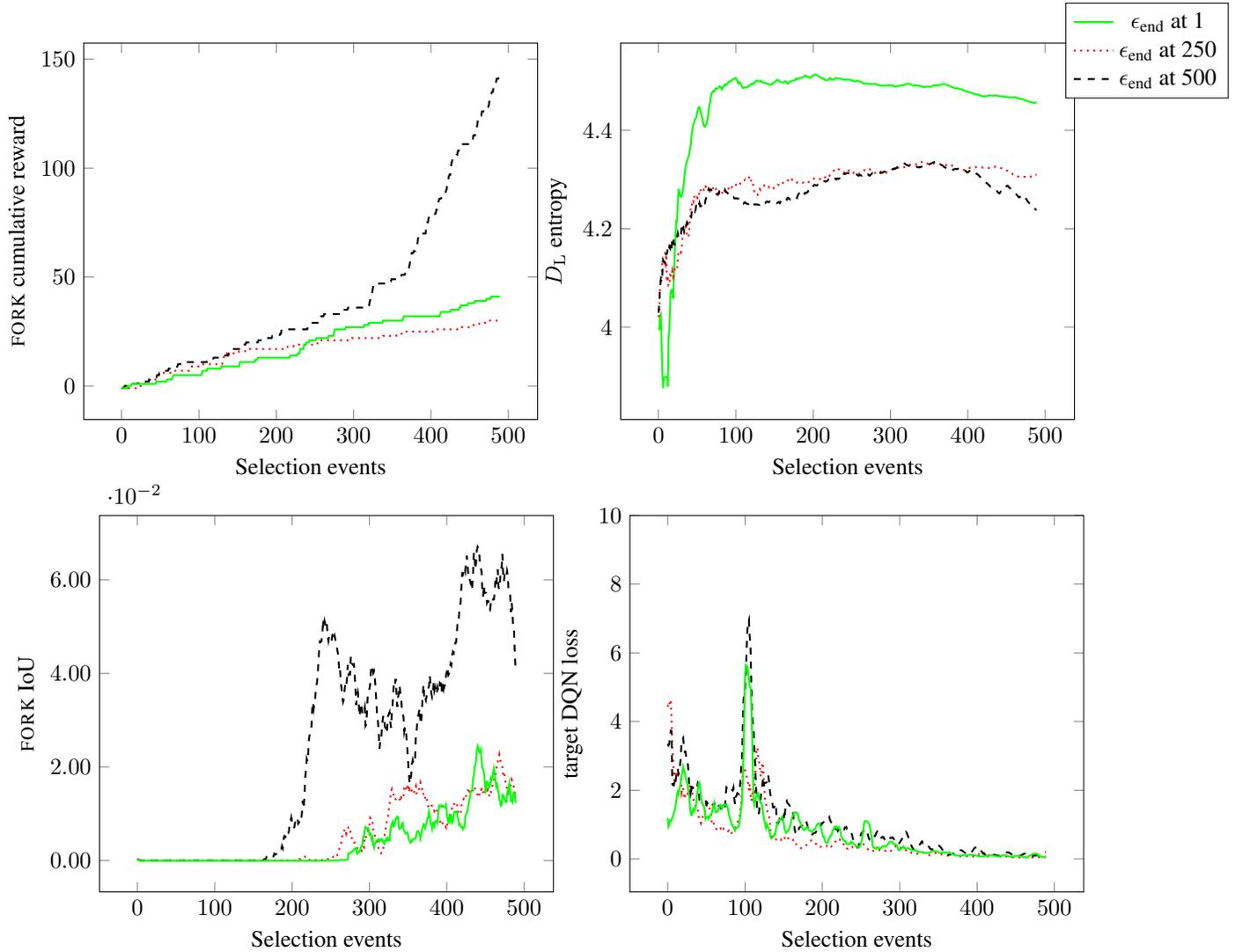
\begin{figure}
        \centering
        \begin{tikzpicture}
            \begin{axis}[
                xlabel={Selection events},
                ylabel={\classname{fork} cumulative reward},
                legend style={anchor=west}
                ]
                \addplot [thick, dotted, color=red, mark size=1pt]
                table [x=step,y=zugw_cr]  {eps_annealing.dat};
                \addplot [thick, dashed, color=black, mark size=1pt]
                table [x=step,y=pawf_cr]  {eps_annealing.dat};
                \addplot [thick, color=green, mark size=1pt]
                table [x=step,y=tpbu_cr]  {eps_annealing.dat};
            \end{axis}
        \end{tikzpicture}%
        \begin{tikzpicture}
            \begin{axis}[
                xlabel={Selection events},
                ylabel={\dlabelled{} entropy},
                legend style={anchor=west},
                ]
                \addplot [thick, color=green, mark size=1pt]
                table [x=step,y=tpbu_entr]  {eps_annealing.dat};
                \addlegendentry{$\epsilon_{\text{end}}$ at 1}
                \addplot [thick, dotted, color=red, mark size=1pt]
                table [x=step,y=zugw_entr]  {eps_annealing.dat};
                \addlegendentry{$\epsilon_{\text{end}}$ at 250}
                \addplot [thick, dashed, color=black, mark size=1pt]
                table [x=step,y=pawf_entr]  {eps_annealing.dat};
                \addlegendentry{$\epsilon_{\text{end}}$ at 500}
            \end{axis}
        \end{tikzpicture}%

        \begin{tikzpicture}
            \begin{axis}[
                xlabel={Selection events},
                ylabel={\classname{fork} IoU},
                legend style={anchor=west},
                ytick={0, 0.02, 0.04, 0.06, 0.08, 0.10},
                y tick label style={
                    /pgf/number format/.cd,
                    fixed,
                    fixed zerofill,
                    precision=2,
                    /tikz/.cd
                }
                ]
                \addplot [thick, dotted, color=red, mark size=1pt]
                table [x=step,y=zugw_smoothed_acc]  {eps_annealing.dat};
                \addplot [thick, dashed, color=black, mark size=1pt]
                table [x=step,y=pawf_smoothed_acc]  {eps_annealing.dat};
                \addplot [thick, color=green, mark size=1pt]
                table [x=step,y=tpbu_smoothed_acc]  {eps_annealing.dat};
            \end{axis}
        \end{tikzpicture}%
        \begin{tikzpicture}
            \begin{axis}[
                xlabel={Selection events},
                ylabel={target \dqn{} loss},
                legend style={anchor=west},
                ymax=10
                ]
                \addplot [thick, dotted, color=red, mark size=1pt]
                table [x=step,y=zugw_smooth_dqn_loss]  {eps_annealing.dat};
                \addplot [thick, dashed, color=black, mark size=1pt]
                table [x=step,y=pawf_smooth_dqn_loss]  {eps_annealing.dat};
                \addplot [thick, color=green, mark size=1pt]
                table [x=step,y=tpbu_smooth_dqn_loss]  {eps_annealing.dat};
            \end{axis}
        \end{tikzpicture}%
        \caption[]{
            Different $\epsilon$-annealing schedules, where the probability of
            picking patches at random decreases from 1 to
            $\epsilon_{\text{end}}$ (here 0.1), lead to different behaviours and
            \ddqn{} performances. Not performing an $\epsilon$-annealing, where
            $\epsilon_{\text{end}}$ is reached at the first step, leads to the
            model directly having to pick patches without having much
            information and trying to exploit lacunary knowledge. This can be
            problematic for relatively rare classes like \classname{fork} since
            finding enough patches to build an effective policy requires
            significant sampling initially. At each sampling event, 64 patches
            are selected from a random selection of 1280 patches, then added to
            \dlabelled{} and one epoch of training is performed on the whole
            \dlabelled{}. The process is repeated until 5\% of \dunlabelled{}
            (\cocotk{}) has been picked and only 250 patches are initially
            randomly sampled and added to \dlabelled{}. The \ddqn{} batch size
            is set to 256 while the buffer contains 6400 experiences. The
            discount factor $\gamma$ is 0.99 while the soft update rate $\beta$
            is 0.002. Rewards are granted when the agent picks
            \classname{fork}-containing patches. Gradients are clipped to 0.01.
            }
            \label{fig:eps_annealing}
        \end{figure}

    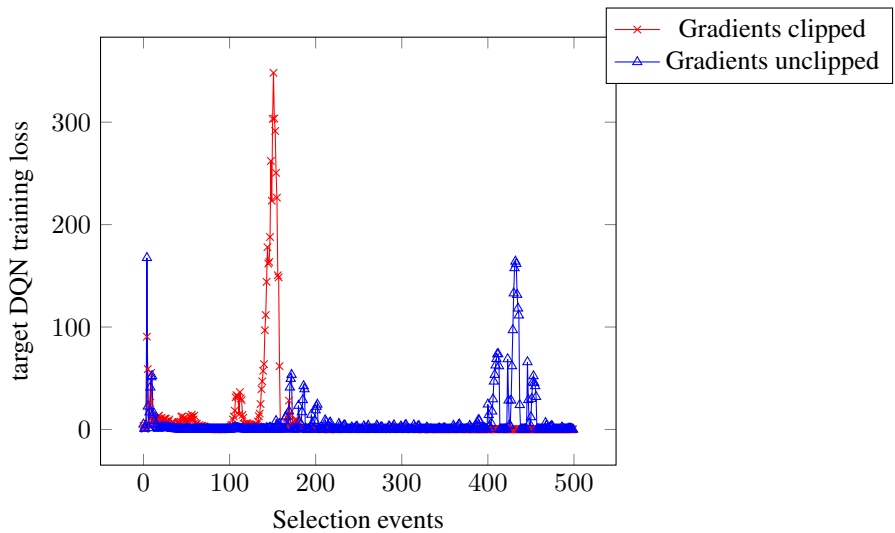
\begin{figure}
        \centering
        \begin{tikzpicture}
            \begin{axis}[
                xlabel={Selection events},
                ylabel={target \dqn{} training loss},
                legend style={anchor=west}
                ]
                \addplot [mark=x, color=red, mark size=2pt]
                table [x=step,y=clipped_yset]  {target_freq_update_grad_clip.dat};
                \addlegendentry{Gradients clipped}
                \addplot [mark=triangle, color=blue, mark size=2pt]
                table [x=step,y=unclipped_enyi] {target_freq_update_grad_clip.dat};
                \addlegendentry{Gradients unclipped}
            \end{axis}
        \end{tikzpicture}%
    \caption[Gradient clipping does not prevent Q-values from reaching high
        levels with a fast soft update.]{Clipping the gradients
        does not prevent the Q-values from reaching high values. Using the soft
        update method from (\ref{eq:soft_update}) is meant to avoid such
        instabilities, however when $\beta$ is set to very high values (here
        $\beta=0.2$), instabilities occur and clipping the gradients to 0.01
        does not prevent the instabilities from happening. At each sampling
        event, 64 patches are selected from a random selection of 1280 patches,
        then added to \dlabelled{} and one epoch of training is performed on the
        whole \dlabelled{}. The process is repeated until 5\% of \dunlabelled{}
        (\cocotk{}) has been picked and only 250 patches are initially randomly
        sampled and added to \dlabelled{}. An $\epsilon$-anneling is applied
        from 1 to 0.1 over 250 steps with $\gamma$ set to 0.99. The \ddqn{}
        batch size is set to 256 while the buffer contains 6400 experiences.
        Rewards are granted when the agent picks \classname{fork}-containing
        patches.
        }
        \label{fig:gradient_clipping}
    \end{figure}

    \begin{figure}
        \centering
        \includegraphics*[width=0.65\textwidth]{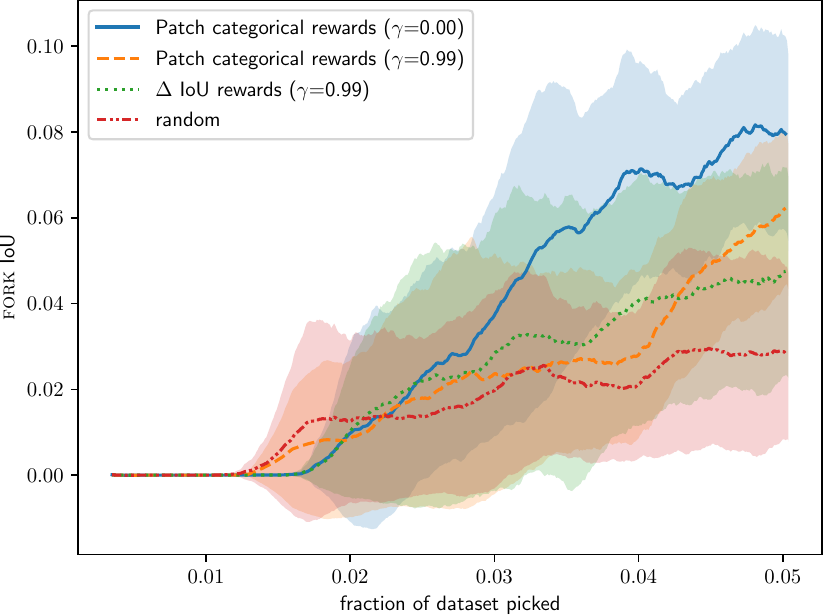}
    \caption[Several policies aiming to maximise the \classname{fork}
        IoU]{Comparison of using categorical rewards associated
        with picking a patch containing the class of interest and using rewards
        linked to the change in IoU of the class of interest ($\Delta$ IoU).
        Using $\gamma=0$ and categorical rewards leads to the fastest
        accumulation of \classname{fork}-containing patches (4x faster than in
        the random policy) followed by the same settings with $\gamma=0$. Using
        $\Delta$ IoU rewards leads to both slow accumulation of
        \classname{fork}-containing patches and slow rise in \classname{fork}
        IoU presumably because the rewards are less granular as all patches
        selected in a step are assigned the same reward. Curves are the means of
        5 seeds with the shading corresponding to their standard deviation, both
        averaged with a 30 step window. At each sampling event, 64 patches are
        selected from a random selection of 1280 patches, then added to
        \dlabelled{} and one epoch of training is performed on the whole
        \dlabelled{}. The process is repeated until 5\% of \dunlabelled{}
        (\cocotk{}) has been picked and only 250 patches are initially randomly
        sampled and added to \dlabelled{}. An $\epsilon$-anneling is applied
        from 1 to 0.1 over 500 steps, which corresponds to 5\% of
        \dunlabelled{}. The \ddqn{} batch size is set to 256 while the buffer
        contains 6400 experiences. The soft update rate $\beta$ is set to 0.02
        and gradients are not clipped.}
        \label{fig:rl_policies}
    \end{figure}

\begin{center}
    \begin{longtable}{rrrrrr}
        \caption{Comparison table of the end accuracies of models trained on 5\%
        of \cocotk{} for each class present in the dataset with the accuracies
        at \dinitial{} corresponding to 2.5\% of \cocotk{} shown for reference.
        Models are initially trained on a \dinitial{} corresponding to 2.5\% of
        the whole dataset then patches are acquired from \dunlabelled{} using
        features calculated from the models trained on \dinitial{} and added to
        \dlabelled{} until 5\% of \cocotk{} has been labelled. Intersection over
        Union figures are given in percentages plus or minus the standard
        deviation over 5 seeds. Outperformance against
        the other policies at 5\% are marked with an * and obtained using the
        Welch t-test for unequal variances with a p-value < 0.05 threshold. The
        largest mean accuracies are highlighted in bold.}
        \label{tab:all_classes_table} \\

        Class &  Random 2.5\% & Random 5\% & Entropy 5\% & BALD 5\% &  \themethod{} 5\% \\
        \hline \Bstrut{}
        \endfirsthead

        Class &  Random 2.5\% & Random 5\% & Entropy 5\% & BALD 5\% &  \themethod{} 5\% \\
        \hline \Bstrut{}
        \endhead

        \hline \multicolumn{6}{r}{{Continued on next page}} \\
        \endfoot

        \hline \hline
        \endlastfoot
        Accuracy/airplane IoU & 36.76$\pm$2.06 & \mbox{40.55$\pm$2.45\hspace{5pt}}&\mbox{33.38$\pm$2.52\hspace{5pt}}&\mbox{36.77$\pm$3.49\hspace{5pt}}&\mbox{\textbf{41.49$\pm$4.91\hspace{5pt}}}\\
        Accuracy/apple IoU & 13.65$\pm$5.74 & \mbox{18.64$\pm$5.57\hspace{5pt}}&\mbox{16.08$\pm$3.25\hspace{5pt}}&\mbox{24.07$\pm$3.36\hspace{5pt}}&\mbox{\textbf{26.31$\pm$2.00\hspace{5pt}}}\\
     Accuracy/backpack IoU &  1.47$\pm$1.38 & \mbox{1.97 $\pm$1.61\hspace{5pt}}&\mbox{1.24 $\pm$1.11\hspace{5pt}}&\mbox{3.73 $\pm$1.97\hspace{5pt}}&\mbox{\textbf{4.79 $\pm$1.47\hspace{5pt}}}\\
       Accuracy/banana IoU & 31.41$\pm$1.63 & \mbox{34.75$\pm$2.75\hspace{5pt}}&\mbox{29.24$\pm$2.02\hspace{5pt}}&\mbox{\textbf{36.67$\pm$1.76\hspace{5pt}}}&\mbox{36.36$\pm$4.03\hspace{5pt}}\\
       Accuracy/banner IoU &  8.92$\pm$1.52 & \mbox{\textbf{10.30$\pm$1.49\hspace{5pt}}}&\mbox{9.41 $\pm$1.87\hspace{5pt}}&\mbox{9.93 $\pm$2.62\hspace{5pt}}&\mbox{8.83 $\pm$0.92\hspace{5pt}}\\
 Accuracy/baseball bat IoU &  0.02$\pm$0.04 & \mbox{0.00 $\pm$0.00\hspace{5pt}}&\mbox{0.15 $\pm$0.34\hspace{5pt}}&\mbox{0.02 $\pm$0.05\hspace{5pt}}&\textbf{5.30 $\pm$1.41}*\\
 Accuracy/baseball glove IoU &  0.00$\pm$0.01 & \mbox{0.55 $\pm$0.81\hspace{5pt}}&\mbox{0.00 $\pm$0.00\hspace{5pt}}&\mbox{0.88 $\pm$1.96\hspace{5pt}}&\textbf{5.64 $\pm$2.63}*\\
         Accuracy/bear IoU & 38.19$\pm$9.74 & \mbox{49.92$\pm$4.40\hspace{5pt}}&\mbox{37.53$\pm$2.79\hspace{5pt}}&\mbox{62.53$\pm$5.25\hspace{5pt}}&\mbox{\textbf{65.02$\pm$2.03\hspace{5pt}}}\\
          Accuracy/bed IoU & 32.98$\pm$1.80 & \mbox{36.86$\pm$2.05\hspace{5pt}}&\mbox{32.60$\pm$1.20\hspace{5pt}}&\mbox{36.81$\pm$1.86\hspace{5pt}}&\mbox{\textbf{38.78$\pm$1.75\hspace{5pt}}}\\
        Accuracy/bench IoU & 14.78$\pm$1.99 & \mbox{18.21$\pm$0.69\hspace{5pt}}&\mbox{14.83$\pm$0.89\hspace{5pt}}&\mbox{20.27$\pm$1.71\hspace{5pt}}&\mbox{\textbf{21.67$\pm$0.97\hspace{5pt}}}\\
      Accuracy/bicycle IoU & 29.23$\pm$4.56 & \mbox{34.80$\pm$3.19\hspace{5pt}}&\mbox{27.91$\pm$5.33\hspace{5pt}}&\mbox{33.67$\pm$2.75\hspace{5pt}}&\textbf{41.56$\pm$2.91}*\\
         Accuracy/bird IoU & 21.72$\pm$4.56 & \mbox{28.50$\pm$4.24\hspace{5pt}}&\mbox{20.49$\pm$8.30\hspace{5pt}}&\mbox{37.24$\pm$4.63\hspace{5pt}}&\mbox{\textbf{41.17$\pm$3.95\hspace{5pt}}}\\
      Accuracy/blanket IoU &  0.55$\pm$0.75 & \mbox{0.79 $\pm$0.55\hspace{5pt}}&\mbox{1.02 $\pm$0.23\hspace{5pt}}&\mbox{0.84 $\pm$0.41\hspace{5pt}}&\mbox{\textbf{2.37 $\pm$1.55\hspace{5pt}}}\\
         Accuracy/boat IoU & 27.63$\pm$2.32 & \mbox{28.86$\pm$0.82\hspace{5pt}}&\mbox{25.09$\pm$2.65\hspace{5pt}}&\mbox{29.87$\pm$0.62\hspace{5pt}}&\textbf{31.84$\pm$1.87}*\\
         Accuracy/book IoU & 12.50$\pm$1.32 & \mbox{14.18$\pm$1.37\hspace{5pt}}&\mbox{11.75$\pm$1.13\hspace{5pt}}&\mbox{17.14$\pm$1.68\hspace{5pt}}&\mbox{\textbf{17.87$\pm$0.92\hspace{5pt}}}\\
       Accuracy/bottle IoU &  9.91$\pm$3.03 & \mbox{12.76$\pm$2.48\hspace{5pt}}&\mbox{10.59$\pm$3.67\hspace{5pt}}&\mbox{16.54$\pm$1.78\hspace{5pt}}&\textbf{19.19$\pm$1.45}*\\
         Accuracy/bowl IoU & 15.03$\pm$1.44 & \mbox{16.88$\pm$1.99\hspace{5pt}}&\mbox{16.59$\pm$2.26\hspace{5pt}}&\mbox{19.06$\pm$0.75\hspace{5pt}}&\textbf{23.35$\pm$1.80}*\\
       Accuracy/branch IoU &  0.40$\pm$0.69 & \mbox{1.48 $\pm$1.67\hspace{5pt}}&\mbox{0.07 $\pm$0.08\hspace{5pt}}&\mbox{0.96 $\pm$0.87\hspace{5pt}}&\mbox{\textbf{2.91 $\pm$1.48\hspace{5pt}}}\\
       Accuracy/bridge IoU &  4.55$\pm$3.34 & \mbox{5.06 $\pm$1.45\hspace{5pt}}&\mbox{3.55 $\pm$1.54\hspace{5pt}}&\mbox{6.15 $\pm$3.37\hspace{5pt}}&\mbox{\textbf{9.80 $\pm$3.15\hspace{5pt}}}\\
     Accuracy/broccoli IoU & 26.37$\pm$5.11 & \mbox{29.75$\pm$5.39\hspace{5pt}}&\mbox{33.15$\pm$3.99\hspace{5pt}}&\mbox{35.22$\pm$1.97\hspace{5pt}}&\mbox{\textbf{36.71$\pm$2.79\hspace{5pt}}}\\
 Accuracy/building-other IoU & 37.66$\pm$0.94 & \mbox{38.98$\pm$0.69\hspace{5pt}}&\mbox{36.71$\pm$0.90\hspace{5pt}}&\mbox{38.49$\pm$0.70\hspace{5pt}}&\mbox{\textbf{39.04$\pm$0.55\hspace{5pt}}}\\
          Accuracy/bus IoU & 45.84$\pm$0.49 & \mbox{50.04$\pm$2.31\hspace{5pt}}&\mbox{44.76$\pm$2.87\hspace{5pt}}&\mbox{45.70$\pm$3.17\hspace{5pt}}&\textbf{57.93$\pm$1.56}*\\
         Accuracy/bush IoU & 10.51$\pm$2.17 & \mbox{11.16$\pm$3.96\hspace{5pt}}&\mbox{10.22$\pm$2.28\hspace{5pt}}&\mbox{8.49 $\pm$0.94\hspace{5pt}}&\mbox{\textbf{14.78$\pm$1.78\hspace{5pt}}}\\
      Accuracy/cabinet IoU &  7.71$\pm$2.25 & \mbox{9.19 $\pm$1.33\hspace{5pt}}&\mbox{11.30$\pm$1.95\hspace{5pt}}&\mbox{11.31$\pm$2.34\hspace{5pt}}&\mbox{\textbf{12.49$\pm$3.80\hspace{5pt}}}\\
         Accuracy/cage IoU &  0.17$\pm$0.26 & \mbox{0.84 $\pm$0.73\hspace{5pt}}&\mbox{0.47 $\pm$0.57\hspace{5pt}}&\mbox{\textbf{3.55 $\pm$0.89\hspace{5pt}}}&\mbox{3.41 $\pm$1.47\hspace{5pt}}\\
         Accuracy/cake IoU & 15.37$\pm$2.38 & \mbox{18.52$\pm$2.95\hspace{5pt}}&\mbox{14.52$\pm$2.04\hspace{5pt}}&\mbox{\textbf{23.28$\pm$1.99\hspace{5pt}}}&\mbox{23.10$\pm$2.29\hspace{5pt}}\\
          Accuracy/car IoU & 25.84$\pm$3.89 & \mbox{30.37$\pm$2.44\hspace{5pt}}&\mbox{28.65$\pm$1.21\hspace{5pt}}&\mbox{31.33$\pm$1.28\hspace{5pt}}&\textbf{35.81$\pm$2.00}*\\
    Accuracy/cardboard IoU &  1.82$\pm$2.40 & \mbox{4.61 $\pm$2.56\hspace{5pt}}&\mbox{1.15 $\pm$1.09\hspace{5pt}}&\mbox{5.05 $\pm$2.37\hspace{5pt}}&\textbf{8.26 $\pm$1.31}*\\
       Accuracy/carpet IoU & 25.63$\pm$2.49 & \mbox{27.28$\pm$2.25\hspace{5pt}}&\mbox{25.38$\pm$3.22\hspace{5pt}}&\mbox{\textbf{28.12$\pm$3.31\hspace{5pt}}}&\mbox{27.00$\pm$2.47\hspace{5pt}}\\
       Accuracy/carrot IoU &  9.77$\pm$4.61 & \mbox{12.50$\pm$5.00\hspace{5pt}}&\mbox{9.03 $\pm$4.95\hspace{5pt}}&\mbox{17.51$\pm$3.81\hspace{5pt}}&\textbf{24.50$\pm$1.05}*\\
          Accuracy/cat IoU & 49.25$\pm$1.66 & \mbox{53.59$\pm$1.46\hspace{5pt}}&\mbox{49.26$\pm$4.30\hspace{5pt}}&\mbox{54.25$\pm$3.10\hspace{5pt}}&\textbf{58.75$\pm$0.57}*\\
 Accuracy/ceiling-other IoU & 37.91$\pm$2.53 & \mbox{40.41$\pm$2.51\hspace{5pt}}&\mbox{38.60$\pm$1.20\hspace{5pt}}&\mbox{38.12$\pm$1.77\hspace{5pt}}&\textbf{43.78$\pm$1.67}*\\
 Accuracy/ceiling-tile IoU &  0.46$\pm$0.69 & \mbox{\textbf{2.12 $\pm$2.93\hspace{5pt}}}&\mbox{0.93 $\pm$1.26\hspace{5pt}}&\mbox{0.00 $\pm$0.00\hspace{5pt}}&\mbox{1.79 $\pm$1.79\hspace{5pt}}\\
   Accuracy/cell phone IoU & 17.72$\pm$6.60 & \mbox{25.32$\pm$3.57\hspace{5pt}}&\mbox{15.53$\pm$3.03\hspace{5pt}}&\mbox{30.42$\pm$1.45\hspace{5pt}}&\mbox{\textbf{31.71$\pm$3.49\hspace{5pt}}}\\
        Accuracy/chair IoU & 11.57$\pm$0.65 & \mbox{14.15$\pm$0.90\hspace{5pt}}&\mbox{12.25$\pm$0.60\hspace{5pt}}&\mbox{13.92$\pm$0.96\hspace{5pt}}&\textbf{17.18$\pm$0.46}*\\
        Accuracy/clock IoU & 35.03$\pm$6.16 & \mbox{36.70$\pm$2.61\hspace{5pt}}&\mbox{35.26$\pm$0.91\hspace{5pt}}&\mbox{42.85$\pm$2.37\hspace{5pt}}&\mbox{\textbf{46.01$\pm$3.31\hspace{5pt}}}\\
        Accuracy/cloth IoU &  0.70$\pm$0.49 & \mbox{0.61 $\pm$0.42\hspace{5pt}}&\mbox{0.77 $\pm$0.41\hspace{5pt}}&\mbox{0.89 $\pm$0.61\hspace{5pt}}&\mbox{\textbf{1.71 $\pm$0.80\hspace{5pt}}}\\
      Accuracy/clothes IoU &  0.34$\pm$0.35 & \mbox{0.39 $\pm$0.54\hspace{5pt}}&\mbox{0.20 $\pm$0.21\hspace{5pt}}&\mbox{\textbf{0.90 $\pm$0.30\hspace{5pt}}}&\mbox{0.90 $\pm$0.73\hspace{5pt}}\\
       Accuracy/clouds IoU & 35.50$\pm$3.45 & \mbox{38.24$\pm$1.29\hspace{5pt}}&\mbox{36.52$\pm$0.71\hspace{5pt}}&\mbox{\textbf{38.78$\pm$1.55\hspace{5pt}}}&\mbox{36.53$\pm$1.19\hspace{5pt}}\\
        Accuracy/couch IoU & 20.62$\pm$2.11 & \mbox{25.54$\pm$1.00\hspace{5pt}}&\mbox{22.99$\pm$0.74\hspace{5pt}}&\mbox{27.17$\pm$1.24\hspace{5pt}}&\textbf{31.03$\pm$1.03}*\\
      Accuracy/counter IoU &  8.21$\pm$1.10 & \mbox{9.49 $\pm$1.45\hspace{5pt}}&\mbox{7.52 $\pm$1.10\hspace{5pt}}&\mbox{10.77$\pm$0.91\hspace{5pt}}&\textbf{12.01$\pm$0.80}*\\
          Accuracy/cow IoU & 26.62$\pm$2.21 & \mbox{33.73$\pm$1.75\hspace{5pt}}&\mbox{25.55$\pm$2.58\hspace{5pt}}&\mbox{32.47$\pm$2.17\hspace{5pt}}&\textbf{42.31$\pm$2.17}*\\
          Accuracy/cup IoU & 13.61$\pm$1.68 & \mbox{16.60$\pm$2.11\hspace{5pt}}&\mbox{14.88$\pm$0.60\hspace{5pt}}&\mbox{18.38$\pm$1.46\hspace{5pt}}&\textbf{22.35$\pm$2.03}*\\
     Accuracy/cupboard IoU &  2.14$\pm$0.63 & \mbox{2.07 $\pm$0.47\hspace{5pt}}&\mbox{2.03 $\pm$0.50\hspace{5pt}}&\mbox{2.33 $\pm$0.44\hspace{5pt}}&\textbf{2.99 $\pm$0.41}*\\
      Accuracy/curtain IoU & 25.06$\pm$2.09 & \mbox{25.02$\pm$1.60\hspace{5pt}}&\mbox{22.93$\pm$1.29\hspace{5pt}}&\mbox{28.27$\pm$3.00\hspace{5pt}}&\textbf{31.63$\pm$1.01}*\\
   Accuracy/desk-stuff IoU & 15.88$\pm$5.89 & \mbox{19.08$\pm$2.70\hspace{5pt}}&\mbox{15.71$\pm$1.20\hspace{5pt}}&\mbox{20.15$\pm$0.68\hspace{5pt}}&\mbox{\textbf{20.22$\pm$3.01\hspace{5pt}}}\\
 Accuracy/dining table IoU & 27.84$\pm$1.14 & \mbox{29.65$\pm$1.07\hspace{5pt}}&\mbox{28.43$\pm$1.44\hspace{5pt}}&\mbox{\textbf{30.65$\pm$1.24\hspace{5pt}}}&\mbox{30.57$\pm$1.10\hspace{5pt}}\\
         Accuracy/dirt IoU & 23.10$\pm$1.40 & \mbox{25.04$\pm$2.28\hspace{5pt}}&\mbox{23.32$\pm$0.55\hspace{5pt}}&\mbox{25.59$\pm$1.14\hspace{5pt}}&\mbox{\textbf{25.66$\pm$1.33\hspace{5pt}}}\\
          Accuracy/dog IoU & 31.39$\pm$3.86 & \mbox{35.85$\pm$2.10\hspace{5pt}}&\mbox{29.77$\pm$2.92\hspace{5pt}}&\mbox{40.65$\pm$2.12\hspace{5pt}}&\textbf{46.68$\pm$2.94}*\\
        Accuracy/donut IoU & 10.76$\pm$4.61 & \mbox{22.40$\pm$4.03\hspace{5pt}}&\mbox{18.05$\pm$7.42\hspace{5pt}}&\mbox{31.87$\pm$2.00\hspace{5pt}}&\mbox{\textbf{33.74$\pm$2.96\hspace{5pt}}}\\
   Accuracy/door-stuff IoU &  7.55$\pm$2.00 & \mbox{8.81 $\pm$2.36\hspace{5pt}}&\mbox{9.32 $\pm$2.05\hspace{5pt}}&\mbox{10.34$\pm$1.85\hspace{5pt}}&\textbf{14.51$\pm$1.43}*\\
     Accuracy/elephant IoU & 62.90$\pm$2.87 & \mbox{69.92$\pm$2.57\hspace{5pt}}&\mbox{62.43$\pm$5.71\hspace{5pt}}&\mbox{70.44$\pm$2.81\hspace{5pt}}&\mbox{\textbf{72.27$\pm$1.52\hspace{5pt}}}\\
        Accuracy/fence IoU & 22.23$\pm$0.84 & \mbox{23.69$\pm$1.36\hspace{5pt}}&\mbox{21.91$\pm$0.86\hspace{5pt}}&\mbox{22.18$\pm$0.55\hspace{5pt}}&\mbox{\textbf{24.27$\pm$0.85\hspace{5pt}}}\\
 Accuracy/fire hydrant IoU & 36.19$\pm$3.30 & \mbox{48.32$\pm$1.73\hspace{5pt}}&\mbox{38.82$\pm$4.21\hspace{5pt}}&\mbox{52.31$\pm$4.77\hspace{5pt}}&\mbox{\textbf{53.39$\pm$3.65\hspace{5pt}}}\\
 Accuracy/floor-marble IoU &  0.07$\pm$0.16 & \mbox{0.06 $\pm$0.07\hspace{5pt}}&\mbox{0.13 $\pm$0.29\hspace{5pt}}&\mbox{0.09 $\pm$0.12\hspace{5pt}}&\mbox{\textbf{0.78 $\pm$0.76\hspace{5pt}}}\\
  Accuracy/floor-other IoU &  7.21$\pm$1.96 & \mbox{8.57 $\pm$0.95\hspace{5pt}}&\mbox{7.76 $\pm$1.44\hspace{5pt}}&\mbox{8.33 $\pm$1.07\hspace{5pt}}&\mbox{\textbf{9.72 $\pm$1.01\hspace{5pt}}}\\
  Accuracy/floor-stone IoU &  0.87$\pm$1.44 & \mbox{0.47 $\pm$0.33\hspace{5pt}}&\mbox{1.42 $\pm$1.74\hspace{5pt}}&\mbox{0.96 $\pm$0.72\hspace{5pt}}&\mbox{\textbf{2.19 $\pm$2.62\hspace{5pt}}}\\
   Accuracy/floor-tile IoU & 27.01$\pm$2.41 & \mbox{30.65$\pm$1.63\hspace{5pt}}&\mbox{26.48$\pm$2.93\hspace{5pt}}&\mbox{30.73$\pm$1.17\hspace{5pt}}&\textbf{34.09$\pm$1.49}*\\
   Accuracy/floor-wood IoU & 27.66$\pm$1.34 & \mbox{30.17$\pm$2.20\hspace{5pt}}&\mbox{27.28$\pm$2.14\hspace{5pt}}&\mbox{30.50$\pm$0.88\hspace{5pt}}&\textbf{33.86$\pm$1.04}*\\
       Accuracy/flower IoU & 13.11$\pm$3.94 & \mbox{13.15$\pm$3.50\hspace{5pt}}&\mbox{15.08$\pm$1.77\hspace{5pt}}&\mbox{15.67$\pm$4.07\hspace{5pt}}&\mbox{\textbf{17.83$\pm$3.72\hspace{5pt}}}\\
          Accuracy/fog IoU &  0.00$\pm$0.00 & \mbox{0.00 $\pm$0.00\hspace{5pt}}&\mbox{0.00 $\pm$0.00\hspace{5pt}}&\mbox{0.00 $\pm$0.00\hspace{5pt}}&\mbox{\textbf{0.90 $\pm$1.48\hspace{5pt}}}\\
   Accuracy/food-other IoU & 11.00$\pm$2.01 & \mbox{11.00$\pm$2.86\hspace{5pt}}&\mbox{11.71$\pm$1.61\hspace{5pt}}&\mbox{15.72$\pm$2.33\hspace{5pt}}&\mbox{\textbf{15.73$\pm$3.23\hspace{5pt}}}\\
         Accuracy/fork IoU &  0.02$\pm$0.04 & \mbox{0.09 $\pm$0.19\hspace{5pt}}&\mbox{0.10 $\pm$0.21\hspace{5pt}}&\mbox{0.46 $\pm$0.47\hspace{5pt}}&\textbf{8.52 $\pm$1.34}*\\
      Accuracy/frisbee IoU & 13.83$\pm$1.21 & \mbox{17.55$\pm$4.16\hspace{5pt}}&\mbox{16.34$\pm$1.48\hspace{5pt}}&\mbox{22.86$\pm$4.01\hspace{5pt}}&\mbox{\textbf{24.51$\pm$2.49\hspace{5pt}}}\\
        Accuracy/fruit IoU &  5.91$\pm$3.66 & \mbox{6.88 $\pm$3.33\hspace{5pt}}&\mbox{2.70 $\pm$1.90\hspace{5pt}}&\mbox{7.76 $\pm$2.38\hspace{5pt}}&\textbf{11.88$\pm$2.45}*\\
 Accuracy/furniture-other IoU &  3.44$\pm$0.50 & \mbox{4.82 $\pm$0.41\hspace{5pt}}&\mbox{4.40 $\pm$0.38\hspace{5pt}}&\mbox{3.81 $\pm$0.91\hspace{5pt}}&\mbox{\textbf{5.21 $\pm$1.05\hspace{5pt}}}\\
      Accuracy/giraffe IoU & 58.36$\pm$3.01 & \mbox{63.92$\pm$1.85\hspace{5pt}}&\mbox{60.09$\pm$3.61\hspace{5pt}}&\mbox{64.29$\pm$0.55\hspace{5pt}}&\textbf{67.67$\pm$1.93}*\\
        Accuracy/grass IoU & 55.90$\pm$0.87 & \mbox{\textbf{58.86$\pm$0.69\hspace{5pt}}}&\mbox{55.42$\pm$1.52\hspace{5pt}}&\mbox{57.43$\pm$0.91\hspace{5pt}}&\mbox{58.25$\pm$1.10\hspace{5pt}}\\
       Accuracy/gravel IoU &  7.83$\pm$0.80 & \mbox{\textbf{7.30 $\pm$0.69\hspace{5pt}}}&\mbox{5.59 $\pm$2.22\hspace{5pt}}&\mbox{5.80 $\pm$0.95\hspace{5pt}}&\mbox{6.46 $\pm$2.18\hspace{5pt}}\\
 Accuracy/ground-other IoU &  3.25$\pm$0.28 & \mbox{3.80 $\pm$0.47\hspace{5pt}}&\mbox{3.24 $\pm$0.71\hspace{5pt}}&\mbox{2.89 $\pm$0.83\hspace{5pt}}&\mbox{\textbf{3.82 $\pm$0.76\hspace{5pt}}}\\
   Accuracy/hair drier IoU &  0.00$\pm$0.00 & \mbox{0.00 $\pm$0.00\hspace{5pt}}&\mbox{0.00 $\pm$0.00\hspace{5pt}}&\mbox{0.00 $\pm$0.00\hspace{5pt}}&\mbox{0.00 $\pm$0.00\hspace{5pt}}\\
      Accuracy/handbag IoU &  0.28$\pm$0.24 & \mbox{0.60 $\pm$0.17\hspace{5pt}}&\mbox{0.47 $\pm$0.38\hspace{5pt}}&\mbox{0.49 $\pm$0.31\hspace{5pt}}&\textbf{1.53 $\pm$0.84}*\\
         Accuracy/hill IoU &  8.99$\pm$1.51 & \mbox{\textbf{9.35 $\pm$0.89\hspace{5pt}}}&\mbox{7.18 $\pm$1.46\hspace{5pt}}&\mbox{9.28 $\pm$0.84\hspace{5pt}}&\mbox{9.02 $\pm$1.15\hspace{5pt}}\\
        Accuracy/horse IoU & 30.17$\pm$3.93 & \mbox{36.97$\pm$3.73\hspace{5pt}}&\mbox{33.47$\pm$2.96\hspace{5pt}}&\mbox{35.01$\pm$2.94\hspace{5pt}}&\textbf{44.53$\pm$4.28}*\\
      Accuracy/hot dog IoU & 14.85$\pm$2.54 & \mbox{21.71$\pm$4.69\hspace{5pt}}&\mbox{17.86$\pm$4.90\hspace{5pt}}&\mbox{23.08$\pm$3.74\hspace{5pt}}&\textbf{28.67$\pm$2.12}*\\
        Accuracy/house IoU &  8.39$\pm$2.52 & \mbox{6.87 $\pm$1.06\hspace{5pt}}&\mbox{8.59 $\pm$1.97\hspace{5pt}}&\mbox{6.43 $\pm$2.20\hspace{5pt}}&\textbf{11.28$\pm$2.41}*\\
     Accuracy/keyboard IoU & 33.86$\pm$3.03 & \mbox{31.90$\pm$3.16\hspace{5pt}}&\mbox{30.97$\pm$4.11\hspace{5pt}}&\mbox{35.82$\pm$2.19\hspace{5pt}}&\mbox{\textbf{38.76$\pm$2.88\hspace{5pt}}}\\
         Accuracy/kite IoU & 16.28$\pm$4.54 & \mbox{18.20$\pm$3.38\hspace{5pt}}&\mbox{15.68$\pm$2.51\hspace{5pt}}&\mbox{17.73$\pm$4.33\hspace{5pt}}&\textbf{23.85$\pm$3.21}*\\
        Accuracy/knife IoU &  0.18$\pm$0.39 & \mbox{0.29 $\pm$0.49\hspace{5pt}}&\mbox{0.18 $\pm$0.39\hspace{5pt}}&\mbox{0.35 $\pm$0.22\hspace{5pt}}&\textbf{4.97 $\pm$0.72}*\\
       Accuracy/laptop IoU & 28.13$\pm$2.03 & \mbox{32.85$\pm$2.16\hspace{5pt}}&\mbox{28.67$\pm$1.85\hspace{5pt}}&\mbox{32.85$\pm$2.66\hspace{5pt}}&\textbf{42.55$\pm$0.97}*\\
       Accuracy/leaves IoU &  4.15$\pm$3.03 & \mbox{\textbf{6.62 $\pm$5.49\hspace{5pt}}}&\mbox{4.38 $\pm$2.23\hspace{5pt}}&\mbox{3.66 $\pm$3.35\hspace{5pt}}&\mbox{4.85 $\pm$2.61\hspace{5pt}}\\
        Accuracy/light IoU &  3.36$\pm$1.75 & \mbox{5.77 $\pm$0.70\hspace{5pt}}&\mbox{2.39 $\pm$1.55\hspace{5pt}}&\mbox{5.24 $\pm$2.57\hspace{5pt}}&\textbf{12.25$\pm$1.09}*\\
          Accuracy/mat IoU &  0.00$\pm$0.00 & \mbox{0.05 $\pm$0.09\hspace{5pt}}&\mbox{0.00 $\pm$0.00\hspace{5pt}}&\mbox{0.10 $\pm$0.19\hspace{5pt}}&\mbox{\textbf{0.51 $\pm$1.14\hspace{5pt}}}\\
        Accuracy/metal IoU &  1.51$\pm$0.92 & \mbox{2.14 $\pm$0.74\hspace{5pt}}&\mbox{1.08 $\pm$0.31\hspace{5pt}}&\mbox{\textbf{2.40 $\pm$1.18\hspace{5pt}}}&\mbox{2.21 $\pm$0.44\hspace{5pt}}\\
    Accuracy/microwave IoU &  5.94$\pm$6.74 & \mbox{10.60$\pm$4.45\hspace{5pt}}&\mbox{4.60 $\pm$4.37\hspace{5pt}}&\mbox{17.99$\pm$2.17\hspace{5pt}}&\mbox{\textbf{21.40$\pm$3.91\hspace{5pt}}}\\
 Accuracy/mirror-stuff IoU &  8.80$\pm$1.64 & \mbox{12.93$\pm$2.12\hspace{5pt}}&\mbox{5.78 $\pm$2.58\hspace{5pt}}&\mbox{10.63$\pm$1.60\hspace{5pt}}&\textbf{17.80$\pm$1.26}*\\
         Accuracy/moss IoU &  0.00$\pm$0.00 & \mbox{0.00 $\pm$0.00\hspace{5pt}}&\mbox{0.00 $\pm$0.00\hspace{5pt}}&\mbox{0.00 $\pm$0.00\hspace{5pt}}&\mbox{0.00 $\pm$0.00\hspace{5pt}}\\
   Accuracy/motorcycle IoU & 48.67$\pm$3.03 & \mbox{52.46$\pm$0.94\hspace{5pt}}&\mbox{50.20$\pm$1.12\hspace{5pt}}&\mbox{49.95$\pm$2.74\hspace{5pt}}&\textbf{57.83$\pm$0.58}*\\
     Accuracy/mountain IoU & 25.52$\pm$3.16 & \mbox{30.27$\pm$3.33\hspace{5pt}}&\mbox{24.76$\pm$5.51\hspace{5pt}}&\mbox{23.27$\pm$2.89\hspace{5pt}}&\mbox{\textbf{32.41$\pm$3.20\hspace{5pt}}}\\
        Accuracy/mouse IoU &  2.57$\pm$3.69 & \mbox{10.97$\pm$10.69\hspace{5pt}}&\mbox{4.83 $\pm$4.14\hspace{5pt}}&\mbox{20.80$\pm$1.07\hspace{5pt}}&\mbox{\textbf{24.42$\pm$4.12\hspace{5pt}}}\\
          Accuracy/mud IoU &  0.63$\pm$1.01 & \mbox{1.51 $\pm$1.89\hspace{5pt}}&\mbox{0.19 $\pm$0.29\hspace{5pt}}&\mbox{0.02 $\pm$0.05\hspace{5pt}}&\mbox{\textbf{2.81 $\pm$0.96\hspace{5pt}}}\\
       Accuracy/napkin IoU &  0.00$\pm$0.00 & \mbox{0.03 $\pm$0.07\hspace{5pt}}&\mbox{0.51 $\pm$0.83\hspace{5pt}}&\mbox{0.09 $\pm$0.18\hspace{5pt}}&\textbf{4.40 $\pm$3.69}*\\
          Accuracy/net IoU & 13.38$\pm$4.80 & \mbox{17.32$\pm$3.16\hspace{5pt}}&\mbox{9.90 $\pm$2.91\hspace{5pt}}&\mbox{11.34$\pm$2.45\hspace{5pt}}&\mbox{\textbf{19.98$\pm$4.45\hspace{5pt}}}\\
       Accuracy/orange IoU & 31.91$\pm$3.90 & \mbox{37.38$\pm$3.33\hspace{5pt}}&\mbox{32.77$\pm$2.83\hspace{5pt}}&\mbox{43.34$\pm$2.67\hspace{5pt}}&\textbf{47.93$\pm$2.58}*\\
         Accuracy/oven IoU & 22.52$\pm$1.30 & \mbox{26.82$\pm$1.14\hspace{5pt}}&\mbox{22.10$\pm$2.31\hspace{5pt}}&\mbox{27.82$\pm$2.07\hspace{5pt}}&\textbf{32.65$\pm$1.79}*\\
        Accuracy/paper IoU &  3.20$\pm$1.07 & \mbox{5.57 $\pm$1.49\hspace{5pt}}&\mbox{3.53 $\pm$1.23\hspace{5pt}}&\mbox{3.74 $\pm$0.61\hspace{5pt}}&\mbox{\textbf{7.20 $\pm$1.35\hspace{5pt}}}\\
 Accuracy/parking meter IoU & 11.61$\pm$3.52 & \mbox{19.63$\pm$2.93\hspace{5pt}}&\mbox{14.09$\pm$4.21\hspace{5pt}}&\mbox{29.42$\pm$4.99\hspace{5pt}}&\mbox{\textbf{30.82$\pm$5.23\hspace{5pt}}}\\
     Accuracy/pavement IoU & 28.06$\pm$1.94 & \mbox{\textbf{27.84$\pm$1.80\hspace{5pt}}}&\mbox{25.95$\pm$2.04\hspace{5pt}}&\mbox{26.03$\pm$1.05\hspace{5pt}}&\mbox{27.06$\pm$1.59\hspace{5pt}}\\
       Accuracy/person IoU & 63.99$\pm$0.90 & \mbox{66.49$\pm$0.41\hspace{5pt}}&\mbox{63.82$\pm$1.29\hspace{5pt}}&\mbox{\textbf{66.77$\pm$0.91\hspace{5pt}}}&\mbox{65.69$\pm$1.35\hspace{5pt}}\\
       Accuracy/pillow IoU &  0.06$\pm$0.14 & \mbox{0.00 $\pm$0.00\hspace{5pt}}&\mbox{0.00 $\pm$0.00\hspace{5pt}}&\mbox{0.00 $\pm$0.00\hspace{5pt}}&\mbox{\textbf{0.00 $\pm$0.01\hspace{5pt}}}\\
        Accuracy/pizza IoU & 42.23$\pm$4.71 & \mbox{46.33$\pm$1.71\hspace{5pt}}&\mbox{44.91$\pm$3.28\hspace{5pt}}&\mbox{48.84$\pm$3.59\hspace{5pt}}&\textbf{53.03$\pm$1.01}*\\
  Accuracy/plant-other IoU & 10.16$\pm$1.07 & \mbox{\textbf{10.73$\pm$0.88\hspace{5pt}}}&\mbox{9.16 $\pm$1.64\hspace{5pt}}&\mbox{10.62$\pm$1.30\hspace{5pt}}&\mbox{10.49$\pm$1.75\hspace{5pt}}\\
      Accuracy/plastic IoU &  0.36$\pm$0.43 & \mbox{0.09 $\pm$0.09\hspace{5pt}}&\mbox{0.28 $\pm$0.24\hspace{5pt}}&\mbox{0.58 $\pm$0.42\hspace{5pt}}&\textbf{1.61 $\pm$0.63}*\\
     Accuracy/platform IoU & 11.32$\pm$2.88 & \mbox{15.32$\pm$1.90\hspace{5pt}}&\mbox{11.32$\pm$3.32\hspace{5pt}}&\mbox{13.29$\pm$1.80\hspace{5pt}}&\textbf{17.78$\pm$0.54}*\\
 Accuracy/playingfield IoU & 56.43$\pm$1.68 & \mbox{55.06$\pm$3.43\hspace{5pt}}&\mbox{53.96$\pm$4.16\hspace{5pt}}&\mbox{\textbf{56.79$\pm$1.01\hspace{5pt}}}&\mbox{56.24$\pm$4.21\hspace{5pt}}\\
 Accuracy/potted plant IoU & 13.02$\pm$1.96 & \mbox{16.01$\pm$1.05\hspace{5pt}}&\mbox{12.84$\pm$1.87\hspace{5pt}}&\mbox{16.48$\pm$2.37\hspace{5pt}}&\mbox{\textbf{16.74$\pm$1.05\hspace{5pt}}}\\
      Accuracy/railing IoU &  1.11$\pm$0.17 & \mbox{1.52 $\pm$0.30\hspace{5pt}}&\mbox{1.02 $\pm$0.16\hspace{5pt}}&\mbox{0.95 $\pm$0.28\hspace{5pt}}&\mbox{\textbf{1.76 $\pm$0.45\hspace{5pt}}}\\
     Accuracy/railroad IoU & 35.85$\pm$5.60 & \mbox{41.37$\pm$0.78\hspace{5pt}}&\mbox{36.99$\pm$2.27\hspace{5pt}}&\mbox{40.32$\pm$0.91\hspace{5pt}}&\mbox{\textbf{42.89$\pm$3.01\hspace{5pt}}}\\
 Accuracy/refrigerator IoU & 18.70$\pm$1.46 & \mbox{22.91$\pm$2.59\hspace{5pt}}&\mbox{21.19$\pm$2.56\hspace{5pt}}&\mbox{29.80$\pm$2.95\hspace{5pt}}&\textbf{42.31$\pm$2.42}*\\
       Accuracy/remote IoU &  5.50$\pm$4.17 & \mbox{8.02 $\pm$4.72\hspace{5pt}}&\mbox{5.54 $\pm$2.90\hspace{5pt}}&\mbox{\textbf{17.46$\pm$3.19\hspace{5pt}}}&\mbox{12.28$\pm$8.52\hspace{5pt}}\\
        Accuracy/river IoU &  8.15$\pm$2.37 & \mbox{10.26$\pm$1.21\hspace{5pt}}&\mbox{10.98$\pm$6.61\hspace{5pt}}&\mbox{9.64 $\pm$4.60\hspace{5pt}}&\mbox{\textbf{13.00$\pm$3.01\hspace{5pt}}}\\
         Accuracy/road IoU & 38.82$\pm$5.27 & \mbox{42.56$\pm$0.49\hspace{5pt}}&\mbox{40.87$\pm$0.62\hspace{5pt}}&\mbox{42.28$\pm$0.49\hspace{5pt}}&\textbf{44.53$\pm$1.97}*\\
         Accuracy/rock IoU & 24.51$\pm$2.86 & \mbox{30.69$\pm$2.70\hspace{5pt}}&\mbox{27.22$\pm$2.47\hspace{5pt}}&\mbox{29.13$\pm$1.58\hspace{5pt}}&\mbox{\textbf{31.59$\pm$1.77\hspace{5pt}}}\\
         Accuracy/roof IoU &  4.28$\pm$0.82 & \mbox{4.96 $\pm$2.16\hspace{5pt}}&\mbox{3.35 $\pm$1.80\hspace{5pt}}&\mbox{3.78 $\pm$1.13\hspace{5pt}}&\mbox{\textbf{6.45 $\pm$1.99\hspace{5pt}}}\\
          Accuracy/rug IoU &  6.97$\pm$0.65 & \mbox{9.69 $\pm$1.03\hspace{5pt}}&\mbox{6.03 $\pm$1.50\hspace{5pt}}&\mbox{8.78 $\pm$1.42\hspace{5pt}}&\textbf{12.55$\pm$1.53}*\\
        Accuracy/salad IoU &  1.54$\pm$1.99 & \mbox{1.26 $\pm$2.00\hspace{5pt}}&\mbox{0.85 $\pm$1.46\hspace{5pt}}&\mbox{0.96 $\pm$0.72\hspace{5pt}}&\textbf{7.90 $\pm$3.53}*\\
         Accuracy/sand IoU & 36.70$\pm$2.05 & \mbox{37.48$\pm$2.51\hspace{5pt}}&\mbox{35.99$\pm$2.42\hspace{5pt}}&\mbox{36.37$\pm$0.86\hspace{5pt}}&\mbox{\textbf{39.66$\pm$0.97\hspace{5pt}}}\\
     Accuracy/sandwich IoU & 17.46$\pm$2.36 & \mbox{18.47$\pm$2.63\hspace{5pt}}&\mbox{16.80$\pm$2.39\hspace{5pt}}&\mbox{19.21$\pm$1.57\hspace{5pt}}&\textbf{23.38$\pm$2.53}*\\
     Accuracy/scissors IoU &  4.13$\pm$5.10 & \mbox{13.34$\pm$2.74\hspace{5pt}}&\mbox{7.08 $\pm$5.32\hspace{5pt}}&\textbf{24.25$\pm$5.72}*&\mbox{16.59$\pm$6.84\hspace{5pt}}\\
          Accuracy/sea IoU & 67.76$\pm$1.34 & \mbox{70.46$\pm$1.68\hspace{5pt}}&\mbox{70.09$\pm$0.80\hspace{5pt}}&\mbox{69.79$\pm$2.20\hspace{5pt}}&\mbox{\textbf{71.60$\pm$2.12\hspace{5pt}}}\\
        Accuracy/sheep IoU & 35.47$\pm$4.03 & \mbox{42.74$\pm$1.89\hspace{5pt}}&\mbox{36.45$\pm$3.94\hspace{5pt}}&\mbox{49.42$\pm$1.74\hspace{5pt}}&\mbox{\textbf{50.12$\pm$3.48\hspace{5pt}}}\\
        Accuracy/shelf IoU &  3.41$\pm$1.64 & \mbox{5.45 $\pm$1.58\hspace{5pt}}&\mbox{2.47 $\pm$1.60\hspace{5pt}}&\mbox{8.79 $\pm$2.44\hspace{5pt}}&\mbox{\textbf{10.42$\pm$1.16\hspace{5pt}}}\\
         Accuracy/sink IoU & 24.56$\pm$3.81 & \mbox{26.56$\pm$3.49\hspace{5pt}}&\mbox{21.32$\pm$4.61\hspace{5pt}}&\mbox{29.87$\pm$3.13\hspace{5pt}}&\textbf{34.75$\pm$1.23}*\\
   Accuracy/skateboard IoU &  3.34$\pm$1.97 & \mbox{6.22 $\pm$1.88\hspace{5pt}}&\mbox{4.93 $\pm$2.53\hspace{5pt}}&\mbox{7.52 $\pm$3.71\hspace{5pt}}&\textbf{16.23$\pm$1.24}*\\
         Accuracy/skis IoU &  1.11$\pm$1.07 & \mbox{1.77 $\pm$1.42\hspace{5pt}}&\mbox{0.91 $\pm$0.92\hspace{5pt}}&\mbox{0.30 $\pm$0.56\hspace{5pt}}&\textbf{10.98$\pm$3.10}*\\
    Accuracy/sky-other IoU & 53.69$\pm$3.02 & \mbox{55.74$\pm$3.18\hspace{5pt}}&\mbox{54.18$\pm$3.03\hspace{5pt}}&\mbox{\textbf{57.92$\pm$1.44\hspace{5pt}}}&\mbox{55.58$\pm$2.93\hspace{5pt}}\\
   Accuracy/skyscraper IoU &  1.65$\pm$2.44 & \mbox{3.60 $\pm$2.15\hspace{5pt}}&\mbox{1.80 $\pm$1.52\hspace{5pt}}&\mbox{1.54 $\pm$2.14\hspace{5pt}}&\textbf{5.81 $\pm$1.03}*\\
         Accuracy/snow IoU & 70.74$\pm$2.55 & \mbox{74.50$\pm$2.45\hspace{5pt}}&\mbox{72.22$\pm$2.23\hspace{5pt}}&\mbox{73.53$\pm$0.96\hspace{5pt}}&\mbox{\textbf{75.61$\pm$1.79\hspace{5pt}}}\\
    Accuracy/snowboard IoU &  1.06$\pm$1.73 & \mbox{3.45 $\pm$1.79\hspace{5pt}}&\mbox{2.77 $\pm$1.68\hspace{5pt}}&\mbox{2.79 $\pm$3.44\hspace{5pt}}&\textbf{13.15$\pm$3.02}*\\
  Accuracy/solid-other IoU &  0.00$\pm$0.00 & \mbox{0.00 $\pm$0.00\hspace{5pt}}&\mbox{0.00 $\pm$0.00\hspace{5pt}}&\mbox{0.00 $\pm$0.00\hspace{5pt}}&\mbox{0.00 $\pm$0.00\hspace{5pt}}\\
        Accuracy/spoon IoU &  0.00$\pm$0.00 & \mbox{0.19 $\pm$0.43\hspace{5pt}}&\mbox{0.09 $\pm$0.20\hspace{5pt}}&\mbox{1.74 $\pm$1.96\hspace{5pt}}&\mbox{\textbf{3.68 $\pm$1.78\hspace{5pt}}}\\
  Accuracy/sports ball IoU &  1.54$\pm$3.28 & \mbox{4.22 $\pm$4.13\hspace{5pt}}&\mbox{1.55 $\pm$2.52\hspace{5pt}}&\mbox{5.61 $\pm$3.92\hspace{5pt}}&\textbf{15.93$\pm$4.82}*\\
       Accuracy/stairs IoU &  1.45$\pm$1.71 & \mbox{1.85 $\pm$0.69\hspace{5pt}}&\mbox{0.79 $\pm$1.08\hspace{5pt}}&\mbox{2.04 $\pm$1.34\hspace{5pt}}&\textbf{6.38 $\pm$1.63}*\\
        Accuracy/stone IoU &  0.82$\pm$0.65 & \mbox{0.79 $\pm$0.60\hspace{5pt}}&\mbox{1.82 $\pm$2.33\hspace{5pt}}&\mbox{\textbf{2.77 $\pm$2.46\hspace{5pt}}}&\mbox{2.31 $\pm$1.13\hspace{5pt}}\\
    Accuracy/stop sign IoU & 55.99$\pm$6.10 & \mbox{\textbf{66.04$\pm$2.88\hspace{5pt}}}&\mbox{56.00$\pm$1.38\hspace{5pt}}&\mbox{63.39$\pm$7.26\hspace{5pt}}&\mbox{63.45$\pm$2.91\hspace{5pt}}\\
        Accuracy/straw IoU &  4.92$\pm$3.25 & \mbox{8.18 $\pm$3.40\hspace{5pt}}&\mbox{8.16 $\pm$5.69\hspace{5pt}}&\mbox{8.46 $\pm$3.37\hspace{5pt}}&\textbf{13.80$\pm$2.30}*\\
 Accuracy/structural-other IoU &  0.46$\pm$0.13 & \mbox{\textbf{0.84 $\pm$0.21\hspace{5pt}}}&\mbox{0.63 $\pm$0.09\hspace{5pt}}&\mbox{0.77 $\pm$0.06\hspace{5pt}}&\mbox{0.64 $\pm$0.16\hspace{5pt}}\\
     Accuracy/suitcase IoU & 15.31$\pm$4.49 & \mbox{23.12$\pm$2.91\hspace{5pt}}&\mbox{17.22$\pm$3.11\hspace{5pt}}&\mbox{25.97$\pm$2.44\hspace{5pt}}&\textbf{28.88$\pm$1.04}*\\
    Accuracy/surfboard IoU & 19.42$\pm$4.73 & \mbox{23.38$\pm$4.33\hspace{5pt}}&\mbox{14.48$\pm$4.69\hspace{5pt}}&\mbox{30.07$\pm$4.94\hspace{5pt}}&\mbox{\textbf{32.52$\pm$4.82\hspace{5pt}}}\\
        Accuracy/table IoU &  8.96$\pm$1.26 & \mbox{9.12 $\pm$0.79\hspace{5pt}}&\mbox{7.67 $\pm$0.96\hspace{5pt}}&\mbox{8.42 $\pm$1.68\hspace{5pt}}&\textbf{11.23$\pm$1.14}*\\
   Accuracy/teddy bear IoU & 37.01$\pm$2.91 & \mbox{42.20$\pm$1.91\hspace{5pt}}&\mbox{37.67$\pm$1.25\hspace{5pt}}&\mbox{48.21$\pm$1.87\hspace{5pt}}&\mbox{\textbf{48.31$\pm$1.11\hspace{5pt}}}\\
 Accuracy/tennis racket IoU & 19.29$\pm$5.04 & \mbox{23.63$\pm$2.90\hspace{5pt}}&\mbox{15.95$\pm$4.95\hspace{5pt}}&\mbox{33.10$\pm$7.09\hspace{5pt}}&\mbox{\textbf{36.49$\pm$3.53\hspace{5pt}}}\\
         Accuracy/tent IoU &  0.33$\pm$0.46 & \mbox{0.51 $\pm$0.68\hspace{5pt}}&\mbox{0.05 $\pm$0.11\hspace{5pt}}&\mbox{0.48 $\pm$0.58\hspace{5pt}}&\mbox{\textbf{0.88 $\pm$0.56\hspace{5pt}}}\\
 Accuracy/textile-other IoU &  2.00$\pm$0.83 & \mbox{2.77 $\pm$0.81\hspace{5pt}}&\mbox{1.96 $\pm$0.54\hspace{5pt}}&\mbox{\textbf{3.46 $\pm$1.11\hspace{5pt}}}&\mbox{3.37 $\pm$0.86\hspace{5pt}}\\
          Accuracy/tie IoU &  0.00$\pm$0.00 & \mbox{0.00 $\pm$0.00\hspace{5pt}}&\mbox{0.00 $\pm$0.00\hspace{5pt}}&\mbox{0.00 $\pm$0.00\hspace{5pt}}&\mbox{\textbf{0.08 $\pm$0.16\hspace{5pt}}}\\
      Accuracy/toaster IoU &  0.00$\pm$0.00 & \mbox{0.00 $\pm$0.00\hspace{5pt}}&\mbox{0.00 $\pm$0.00\hspace{5pt}}&\mbox{0.01 $\pm$0.02\hspace{5pt}}&\mbox{\textbf{0.73 $\pm$1.13\hspace{5pt}}}\\
       Accuracy/toilet IoU & 38.10$\pm$2.73 & \mbox{43.13$\pm$2.51\hspace{5pt}}&\mbox{38.22$\pm$3.63\hspace{5pt}}&\mbox{45.48$\pm$1.86\hspace{5pt}}&\mbox{\textbf{47.52$\pm$3.01\hspace{5pt}}}\\
   Accuracy/toothbrush IoU &  0.00$\pm$0.00 & \mbox{\textbf{0.59 $\pm$1.33\hspace{5pt}}}&\mbox{0.05 $\pm$0.10\hspace{5pt}}&\mbox{0.05 $\pm$0.12\hspace{5pt}}&\mbox{0.28 $\pm$0.38\hspace{5pt}}\\
        Accuracy/towel IoU &  0.11$\pm$0.19 & \mbox{0.94 $\pm$1.07\hspace{5pt}}&\mbox{0.26 $\pm$0.30\hspace{5pt}}&\mbox{0.89 $\pm$0.61\hspace{5pt}}&\textbf{8.48 $\pm$1.78}*\\
 Accuracy/traffic light IoU & 20.46$\pm$2.13 & \mbox{25.43$\pm$2.45\hspace{5pt}}&\mbox{11.65$\pm$6.75\hspace{5pt}}&\mbox{22.32$\pm$3.58\hspace{5pt}}&\textbf{31.61$\pm$1.77}*\\
        Accuracy/train IoU & 41.74$\pm$1.10 & \mbox{47.41$\pm$1.57\hspace{5pt}}&\mbox{40.88$\pm$3.22\hspace{5pt}}&\mbox{44.40$\pm$1.55\hspace{5pt}}&\textbf{51.82$\pm$2.72}*\\
         Accuracy/tree IoU & 58.77$\pm$0.34 & \mbox{\textbf{60.50$\pm$0.83\hspace{5pt}}}&\mbox{59.03$\pm$0.71\hspace{5pt}}&\mbox{59.45$\pm$0.41\hspace{5pt}}&\mbox{59.77$\pm$0.42\hspace{5pt}}\\
        Accuracy/truck IoU & 19.17$\pm$1.34 & \mbox{21.13$\pm$2.65\hspace{5pt}}&\mbox{19.39$\pm$2.09\hspace{5pt}}&\mbox{22.19$\pm$1.92\hspace{5pt}}&\textbf{27.45$\pm$4.33}*\\
           Accuracy/tv IoU & 26.38$\pm$2.48 & \mbox{29.62$\pm$1.58\hspace{5pt}}&\mbox{26.50$\pm$1.39\hspace{5pt}}&\mbox{31.49$\pm$1.72\hspace{5pt}}&\textbf{37.79$\pm$3.11}*\\
     Accuracy/umbrella IoU & 28.20$\pm$3.45 & \mbox{30.73$\pm$1.35\hspace{5pt}}&\mbox{26.20$\pm$0.96\hspace{5pt}}&\mbox{37.57$\pm$2.13\hspace{5pt}}&\mbox{\textbf{39.11$\pm$1.16\hspace{5pt}}}\\
         Accuracy/vase IoU & 15.47$\pm$3.79 & \mbox{21.86$\pm$2.93\hspace{5pt}}&\mbox{15.15$\pm$3.37\hspace{5pt}}&\mbox{\textbf{24.68$\pm$2.48\hspace{5pt}}}&\mbox{24.48$\pm$3.30\hspace{5pt}}\\
    Accuracy/vegetable IoU &  7.36$\pm$4.46 & \mbox{8.98 $\pm$1.43\hspace{5pt}}&\mbox{8.43 $\pm$3.12\hspace{5pt}}&\mbox{\textbf{14.96$\pm$2.06\hspace{5pt}}}&\mbox{13.82$\pm$1.80\hspace{5pt}}\\
   Accuracy/wall-brick IoU & 18.43$\pm$1.40 & \mbox{19.53$\pm$2.37\hspace{5pt}}&\mbox{18.01$\pm$1.54\hspace{5pt}}&\mbox{19.50$\pm$1.37\hspace{5pt}}&\mbox{\textbf{20.20$\pm$1.30\hspace{5pt}}}\\
 Accuracy/wall-concrete IoU &  1.50$\pm$0.37 & \mbox{2.01 $\pm$0.68\hspace{5pt}}&\mbox{1.52 $\pm$0.54\hspace{5pt}}&\mbox{1.27 $\pm$0.56\hspace{5pt}}&\mbox{\textbf{2.41 $\pm$0.67\hspace{5pt}}}\\
   Accuracy/wall-other IoU &  8.58$\pm$0.27 & \mbox{9.01 $\pm$0.49\hspace{5pt}}&\mbox{8.88 $\pm$0.40\hspace{5pt}}&\mbox{9.05 $\pm$0.30\hspace{5pt}}&\mbox{\textbf{9.28 $\pm$0.17\hspace{5pt}}}\\
   Accuracy/wall-panel IoU &  0.21$\pm$0.14 & \mbox{\textbf{0.27 $\pm$0.46\hspace{5pt}}}&\mbox{0.25 $\pm$0.35\hspace{5pt}}&\mbox{0.12 $\pm$0.16\hspace{5pt}}&\mbox{0.18 $\pm$0.20\hspace{5pt}}\\
   Accuracy/wall-stone IoU & 10.13$\pm$2.31 & \mbox{9.61 $\pm$2.85\hspace{5pt}}&\mbox{10.52$\pm$2.03\hspace{5pt}}&\mbox{9.48 $\pm$3.22\hspace{5pt}}&\mbox{\textbf{12.34$\pm$1.88\hspace{5pt}}}\\
    Accuracy/wall-tile IoU & 33.72$\pm$1.46 & \mbox{34.04$\pm$2.31\hspace{5pt}}&\mbox{32.78$\pm$1.86\hspace{5pt}}&\mbox{34.61$\pm$1.35\hspace{5pt}}&\textbf{36.84$\pm$1.35}*\\
    Accuracy/wall-wood IoU & 10.56$\pm$2.04 & \mbox{13.73$\pm$2.02\hspace{5pt}}&\mbox{12.13$\pm$2.03\hspace{5pt}}&\mbox{13.84$\pm$1.68\hspace{5pt}}&\mbox{\textbf{14.29$\pm$2.11\hspace{5pt}}}\\
  Accuracy/water-other IoU & 14.34$\pm$2.75 & \mbox{14.02$\pm$1.52\hspace{5pt}}&\mbox{14.19$\pm$2.50\hspace{5pt}}&\mbox{\textbf{15.52$\pm$2.81\hspace{5pt}}}&\mbox{14.44$\pm$2.86\hspace{5pt}}\\
   Accuracy/waterdrops IoU &  0.01$\pm$0.01 & \mbox{0.06 $\pm$0.13\hspace{5pt}}&\mbox{0.00 $\pm$0.00\hspace{5pt}}&\mbox{0.00 $\pm$0.00\hspace{5pt}}&\mbox{\textbf{0.81 $\pm$1.00\hspace{5pt}}}\\
 Accuracy/window-blind IoU & 15.34$\pm$1.79 & \mbox{20.84$\pm$2.12\hspace{5pt}}&\mbox{16.24$\pm$2.33\hspace{5pt}}&\mbox{20.94$\pm$1.01\hspace{5pt}}&\textbf{23.17$\pm$1.48}*\\
 Accuracy/window-other IoU & 19.47$\pm$1.65 & \mbox{21.83$\pm$0.85\hspace{5pt}}&\mbox{19.48$\pm$0.60\hspace{5pt}}&\mbox{21.75$\pm$0.84\hspace{5pt}}&\textbf{23.66$\pm$1.21}*\\
   Accuracy/wine glass IoU & 11.83$\pm$6.35 & \mbox{18.58$\pm$4.20\hspace{5pt}}&\mbox{10.70$\pm$6.24\hspace{5pt}}&\mbox{21.00$\pm$4.17\hspace{5pt}}&\mbox{\textbf{24.62$\pm$1.98\hspace{5pt}}}\\
         Accuracy/wood IoU &  3.55$\pm$0.56 & \mbox{4.45 $\pm$1.55\hspace{5pt}}&\mbox{3.49 $\pm$0.92\hspace{5pt}}&\mbox{4.44 $\pm$1.23\hspace{5pt}}&\mbox{\textbf{5.31 $\pm$0.42\hspace{5pt}}}\\
        Accuracy/zebra IoU & 74.49$\pm$1.17 & \mbox{74.77$\pm$2.92\hspace{5pt}}&\mbox{73.50$\pm$2.68\hspace{5pt}}&\mbox{\textbf{75.75$\pm$1.65\hspace{5pt}}}&\mbox{74.64$\pm$1.73\hspace{5pt}}\\
    \end{longtable}
\end{center}

\end{document}